\documentclass[letterpaper, 11pt]{article}  
\usepackage[a4paper, total={6.5in, 9in}]{geometry}

\usepackage{cite}
\usepackage{amsmath,amssymb,amsfonts}
\usepackage{algorithm2e}
\usepackage{graphicx}
\usepackage{textcomp}
\usepackage{xcolor}
\usepackage{booktabs}
\usepackage{placeins}
\usepackage{tcolorbox}
\usepackage{svg}
\usepackage{hyperref}
\usepackage{multirow}
\usepackage{amsthm}
\newtheorem{example}{Example}
\newtheorem{theorem}{Theorem}
\newtheorem{definition}{Definition}

\title{\LARGE \bf
Pointwise-in-Time Explanation for Linear Temporal Logic Rules
\thanks{This research was funded in part by a National Defense Science and Engineering Graduate Fellowship.}
}

\author{Noel Brindise and Cédric Langbort
\thanks{N. Brindise and C. Langbort are with the Department of Aerospace Engineering, Grainger College of Engineering, University of Illinois Urbana-Champaign,
         104 S Wright St, Urbana, IL 61801, USA
        {\tt\small nbrindi2@illinois.edu, langbort@illinois.edu}}
}

\begin{document}

\maketitle
\thispagestyle{empty}
\pagestyle{empty}

\begin{abstract}
The new field of Explainable Planning (XAIP) has produced a variety of approaches to explain and describe the behavior of autonomous agents to human observers. Many summarize agent behavior in terms of the constraints, or ``rules,'' which the agent adheres to during its trajectories. In this work, we narrow the focus from summary to specific moments in individual trajectories, offering a ``pointwise-in-time'' view. Our novel framework, which we define on Linear Temporal Logic (LTL) rules, assigns an intuitive \textit{status} to any rule in order to describe the trajectory progress at individual time steps; here, a rule is classified as \textit{active}, \textit{satisfied}, \textit{inactive}, or \textit{violated}. Given a trajectory, a user may query for status of specific LTL rules at individual trajectory time steps.
\par In this paper, we present this novel framework, named Rule Status Assessment (RSA), and provide an example of its implementation. We find that pointwise-in-time status assessment is useful as a post-hoc diagnostic, enabling a user to systematically track the agent's behavior with respect to a set of rules.
\par This draft has been updated as of 11 June 2025 to improve clarity of proofs and provide corrections.
\end{abstract}

\section{Introduction}\label{intro}

Planning and performing on their own, autonomous agents are useful in a wide variety of roles, from self-driving vehicles \cite{xing2021toward} and task-executing robots \cite{seeber2020collaborating} to decision-making software \cite{franklin1997agent}. However, it can be challenging to understand or characterize agent behavior. The emerging field of Explainable Planning (XAIP), a subfield of Explainable AI (xAI), takes on this explanatory challenge with mixed results \cite{sreedharan2022explainable, lindsay2020using, ijcai2020-669}. Some approaches allow the user to query hypotheticals (``What would [agent] do in [state]?'', etc.) \cite{boggess2022toward}; others generate domain alterations that would cause hypothetical route changes (``What needs to change to make [plan] optimal instead?'') \cite{brandao2022not} or use visualizations to help reconcile a human's mental model with that of the agent (``Where does the agent's behavior differ from my expectations?'') \cite{kumar2022vizxp}. 
\par A large body of work explains agent plans as a whole using \textit{rule inference}, an approach originating in formal methods and software model checking \cite{neider2018learning}. Here, an agent is observed, and the explainer infers which ``rules'' or constraints the agent tends to follow (``What does the agent usually, or never, do?''). Rules may be expressed in signal temporal logic (STL) \cite{bombara2016decision, bombara2021offline} or, more commonly, linear temporal logic (LTL) \cite{lemieux2015general, chou2020explaining}. Approaches have learned contrasts between acceptable and unsatisfactory plans \cite{kim2019bayesian}, inferred rules from purely positive behavior \cite{roy2023learning} and noisy data \cite{gaglione2021learning}. In general, rule inference provides global summaries of agent behavior by providing a set of satisfied (or violated) rules. \par For live assessment, the related field of LTL \textit{monitoring} predicts the future satisfaction of rules at a current point in a(n unfinished) trajectory \cite{bartocci2018counting}; similar methods have been used for STL rules in dynamic planning \cite{aasi2021inferring}.
\par While global summaries produced by inference and monitoring are certainly valuable, they do not address local behavior of the agent, which evokes other questions (``Which rules is the agent following at time $t$? Which rules are not relevant at $t$?''). We may ask, for example, whether a rule is still in progress at $t$ or already completed. Moreover, implication-type ($x\rightarrow y$) rules require a consequence $y$ only after a triggering condition $x$ occurs, making the rule trivially satisfied when $x$ does not occur; we may want to know which rules are thus ``triggered'' at a given $t$.
\par In our work, we pursue an aspect of local, time step dependent diagnostics which, to the best of the authors' knowledge, remains unaddressed. We consider trajectories of an arbitrary, black-box agent and define a notion of rule \textit{status}, which we use to assess when and how a system trajectory progresses through each behavior (LTL rule) of interest. This is accomplished by defining notions of \textit{active}, \textit{inactive}, \textit{satisfied}, and \textit{violated} rule status, such that rules are uniquely assigned a status dependent on trajectory and time step. We then introduce an algorithm to perform rule status assessment. We categorize this novel time step-dependent analysis method as``pointwise-in-time'' explanation and demonstrate its potential applications on illustrative examples.
\par

\section{Background}\label{background}
\par We consider discrete-time, discrete-state systems, a common setting for modeling autonomous agents. Such systems include Markov decision processes (MDPs) and Kripke structures; we define both below. We then describe Linear Temporal Logic (LTL), a temporal-logical language which can express constraints (``rules'') on such systems.

\subsection{System Setting}\label{kripkebackground} Fundamentally, our Rule Status Assessment may be applied to any dynamical system which produces a sequence of `labels' or `atomic propositions.' One way to express such a system is via a Kripke structure as in Definition \ref{Kripkedef}.
\begin{definition}[Kripke structure]\label{Kripkedef}
Given a set $P$ of labels, the Kripke structure $K$ over $P$ is a tuple 
\begin{equation}
    K=(S,I,\mathcal{T},\mathcal{L}).
\end{equation}
where $S$ is a finite set of states, $I\subseteq S$ is the set of initial states, and $\mathcal{T}\subseteq S \times S$ is a transition relation with $(\sigma,\sigma')\in \mathcal{T},$ $\sigma'\in S$ for all $\sigma\in S$. Finally, $\mathcal{L} : S \rightarrow 2^{P}$ is a labeling function assigning a set $L\subseteq P$ to each state in $S$. 
\end{definition}
For clarity, we establish separate notation for a state variable $s_t$ and a constant state $\sigma\in S$, i.e., the expression $s_2=\sigma_1$ signifies \textit{the state at time $t=2$ is} $\sigma_1$. Each unique label is denoted $\alpha\in P$.
\par Now, such a system acting over a finite discrete time interval $\{t\ |\ T_0\leq t\leq T_f\}$, where $T_0,t,T_f\in\mathbb{N}$, produces a run:

\begin{definition}[Run] A \textit{run} $\kappa = s_{T_0},...,s_{T_f}$ of the system must satisfy $s_{T_0}\in I$ and $(s_t,s_{t+1})\in R$ for all $t\in\{T_0,...,T_f\}$.
\end{definition}

Importantly, each $\kappa$ induces a \textit{trace} $\rho$:
\begin{definition}[Finite Trace]
For any state $s_t\in S$, $T_0\leq t \leq T_f$, let $\mathcal{L}(s_t) = L_t$, where $L_t$ is a set of labels. A trace is the sequence $\rho = (L_{T_0},...,L_{T_f})$ produced by trajectory $(s_{T_0},...,s_{T_f})$ from time step $T_0$ to $T_f$.
\end{definition}
We may further break traces into \textit{segments:}
\begin{definition}[Trace Segment]
    The trace segment $\rho^{t...}=(L_t,...,L_{T_f})$ is the part of the full trace $\rho$ beginning at $t$ and ending at $T_f$ $(T_0\leq t \leq T_f)$.
\end{definition}
A segment from a time $t$ to the final time in the (finite) trace $T_f$ may also be called a \textit{suffix}.

\par Particularly for Reinforcement Learning applications, the setting may alternatively be expressed as a labeled Markov Decision Process:

\begin{definition}[Labeled Markov Decision Process]
    A labeled MDP is a tuple
    \begin{equation}
        \mathcal{M}=\langle S, A, T, R, P, \mathcal{L}\rangle,
    \end{equation}
    where $S$ is a finite set of discrete states, $A$ is a finite set of actions, $T: S\times A \times S \rightarrow \mathbb{R}$ is a stochastic transition function, and $R: S\times A \rightarrow \mathbb{R}$ is a reward function. $P$ is a finite set of labels, and $\mathcal{L} : S \rightarrow 2^P$ is the labeling function as in Definition \ref{Kripkedef}.
\end{definition}

Here, $\mathcal{M}$ produces a trace according to $\mathcal{L}$ similarly to $K$.

\subsection{Linear Temporal Logic}\label{ltlbackground} Linear temporal logic (LTL) is a human-intuitive way to express time- and order-dependent specifications \cite{6702421}, \cite{camacho2019learning}. An LTL formula $\varphi$ may be evaluated on a trace $\rho$ provided $\varphi$ is defined on labels in $P$.

\begin{definition}[Linear Temporal Logic Formula]\label{LTLdef}
For the set of LTL formulas over a finite set $P$ of labels,
 \par $\bullet$ if $\alpha\in P$, then $\alpha$ is itself an LTL formula.
    \par $\bullet$ if $\varphi_1$ and $\varphi_2$ are LTL formulas, then $\neg \varphi_1$, $\varphi_1\vee\varphi_2$, $\mathbf{X}\varphi_1$, and $\varphi_1\mathbf{U}\varphi_2$ are LTL formulas.
\end{definition}
\begin{definition}[Formula Evaluation on Finite Traces]\label{FiniteTraceDef}
    LTL formula $\varphi$ is true on finite trace $\rho^{t'...} = (L_{t'},...,L_{T_f})$, denoted $\rho^{t'...}\models\varphi$, if $T_0\leq t' \leq T_f$ \textbf{and}:
        \par$\bullet$ $\rho^{t'...} \models \alpha$ where $\alpha\in P\quad$ iff $\quad\alpha\in L_0$
        \par$\bullet$ $\rho^{t'...}\models \neg \varphi\quad$ iff $\quad\rho^{t'...}\not\models\varphi$
        \par$\bullet$ $\rho^{t'...} \models \varphi_1 \wedge \varphi_2\quad$ iff $\quad\rho^{t'...}\models\varphi_1$ and $\rho^{t'...}\models\varphi_2$
        \par and for the temporal operators \textup{Next} $\mathbf{X}$ and \textup{Until} $\mathbf{U}$, 
    \par$\bullet$ $\rho^{t'...} \models \mathbf{X}\varphi\quad$ iff $\quad \rho^{t'+1...}\models\varphi$ \textbf{and} $t'<T_f$
    \par$\bullet$ $\rho^{t'...}\models\varphi_1\mathbf{U}\varphi_2\quad$ iff $\quad \exists i\geq t'$ s.t. $\rho^{i...}\models\varphi_2$ and $\rho^{k...} \models \varphi_1$ for all $k$ s.t. $t'\leq k < i$ \textbf{and} $i\leq T_f$
\end{definition}
Table \ref{tab:LTLoperators} intuitively describes LTL operators. Note that all of the operators listed can be constructed from $\mathbf{X,U},\vee,$ and $\neg$ \cite{pnueli1977temporal}. Lastly, we define formula \textit{arguments}, so that we may decompose formulas into their constituent parts.

\begin{definition}[Arguments of an LTL formula]\label{def:formula_args}
Consider the LTL order of operations: (1) grouping symbols; (2) $\neg, \mathbf{X}$, and other unary operators; (3) $\mathbf{U}$ and other temporal binary operators; and (4) $\vee, \wedge, \rightarrow$. For LTL formula $\varphi$, all $\varphi_j$ bound by the weakest operator are the \textit{arguments} of $\varphi$.
\end{definition}
For example, $\varphi=\mathbf{G}\alpha_1\vee\alpha_2$ has arguments $\varphi_1 = \mathbf{G}\alpha_1$, $\varphi_2 = \alpha_2$. For labels, the argument is the formula itself ($\alpha$).

\begin{table}[t]
\centering
\begin{tabular}{cp{14 cm}}
\toprule
 \textbf{Operator}&\textbf{Meaning}\\
 \midrule
\textbf{G}$\varphi_1$&\textbf{Global}: $\varphi_1$ is always true.\\
\midrule
\textbf{F}$\varphi_1$&\textbf{Eventual}: $\varphi_1$ eventually occurs.\\
\midrule
\textbf{X}$\varphi_1$&\textbf{Next}: $\varphi_1$ must occur at the next time step.\\
\midrule
$\varphi_1$\textbf{U}$\varphi_2$&\textbf{Until}: $\varphi_1$ remains true until $\varphi_2$ occurs (and $\varphi_2$ must occur)\\
\midrule
$\varphi_1$\textbf{W}$\varphi_2$&\textbf{Weak until}: $\varphi_1$ must remain true (1) always, or (2) until $\varphi_2$ becomes true.\\
\midrule
$\varphi_1$\textbf{R}$\varphi_2$&\textbf{Release}: $\varphi_2$ must remain true (1) always, or (2) until (and including) the time step when $\varphi_1$ becomes true.\\
\midrule
$\varphi_1$\textbf{M}$\varphi_2$&\textbf{Strong release}: True only if Condition (2) for Release is met.\\
\midrule
$\varphi_1\wedge\varphi_2$&\textbf{Conjunction}: both $\varphi_1$ and $\varphi_2$ must be true\\
\midrule
$\varphi_1\vee\varphi_2$&\textbf{Disjunction}: $\varphi_1$ or $\varphi_2$ or both must be true\\
\midrule
$\varphi_1\rightarrow\varphi_2$&\textbf{Implication}: if $\varphi_1$ is true, $\varphi_2$ must also be true\\
\bottomrule\\
\end{tabular}
\caption{\label{tab:LTLoperators} Selected LTL operators.}
\end{table}

\section{Problem Statement}\label{problemstatement}
We consider a scenario in which an agent with unknown policy (a ``black box'' agent) is observed by a human. We suppose the human wishes to characterize the agent's behavior in terms of a set of rules; these rules may represent tasks or safety constraints, for example. \textit{A priori}, an agent trajectory may or may not follow any of the rules. 
\par We suppose the human would like to know (1) which rules a trajectory follows, as well as (2) when and how the trajectory progresses through each rule. For example, (2) is relevant when a rule has subtasks (``do this, then that'') or multiple fulfillment conditions (``do this or that''). The second question requires diagnostics which check for specific conditions at each time step in the trace; here, rule inference, which examines entire traces at once, is insufficient.
\subsection{Formal Problem Setting}
We consider agents whose dynamics are expressible as a Kripke structure, as well as a set of LTL rules over $P$ which describe behaviors of interest. Given an observed trace $\rho$, our goal is to assess the progress of rules at any time step; we propose the assignment of a \textit{status} to each rule and its arguments at query points $t^*\in \{T_0,...,T_f\}$ for all trace suffixes $\rho^{t^*_0...}$ where $t^*_0\leq t^*$. 

\par Moving towards a notion of rule status, we begin with two observations. Firstly, consider a trace $\rho$ and rule $\varphi$. Suppose some $t_0$ exists such that $\varphi$ is true on $\rho^{t_0...}$ no matter which labels are contained in any $L_{t\geq t_0}$. In this case, the trace beyond $t_0$ appears to be arbitrary with respect to $\varphi$:
\begin{definition}[Arbitrariness of suffix]\label{arbitraryDef}
For LTL formula $\varphi$, its argument(s) $\varphi_j$, trace $\rho^{T_0...}$, and associated segment $\rho^{t_0...}$ where $T_0\leq t_0 \leq T_f$, we say that $\rho^{t'...}$ is \textup{arbitrary} with respect to $\varphi$ if $\rho^{t_0...}\models\varphi$ is independent of the truth of $\rho^{t...}\models\varphi_j$ for all $t$ in $t'\leq t\leq T_f$ for some $t_0\leq t' \leq T_f$. 
\end{definition}
Secondly, we consider implication rules $\varphi = \varphi_1\rightarrow\varphi_2$. When $\varphi$ is true on $\rho$, there are two very distinct possibilities: either $\varphi_1$ has occurred in $\rho$, in which case $\varphi_2$ must be made true; or $\varphi_1$ did not occur, in which case $\varphi_2$ plays no role. We call such a $\varphi_1$ a \textit{precondition}:
\begin{definition}[Precondition]\label{preconditionDef}
The \textup{precondition} of an LTL formula $\varphi$ is defined as
\begin{enumerate}
    \item $\varphi_1$ if $\varphi$ has the form $\varphi_1\rightarrow\varphi_2$
    \item $\top$ otherwise
\end{enumerate} 
\end{definition}
Motivated by these observations, we characterize the roles of rules at time steps by defining a novel LTL rule status.
\begin{definition}[Status of LTL formula]\label{activedef}
We define the following statuses for an LTL formula $\varphi$ given a trace $\rho^{T_0...}$ and times $t_0,t\in\{T_0,...,T_f\}$, $t_0\leq t$:
\par $\bullet$ $\varphi$ is \textup{active (a)} at $t$ iff (1) $\rho^{t_0...}\models\varphi$, (2) $\rho^{t_0...}\models\psi$ for precondition $\psi$ of $\varphi$, and (3) $\rho^{t...}$ is not arbitrary. 

\par $\bullet$ $\varphi$ is \textup{satisfied (s)} at $t$ iff (1) $\varphi$ \textup{active} at $t$ and (2) $t=T_f$ or $\varphi$ not active at $t+1$.
\par $\bullet$ $\varphi$ is \textup{inactive (i)} at $t$ iff (1) $\rho^{t_0...}\models\varphi$ and (2) $\varphi$ is not \textup{active} at $t$. 
\par $\bullet$ $\varphi$ is \textup{violated (v)} at $t$ iff $\rho^{t_0...}\not\models\varphi$.
\end{definition}
\par We briefly illustrate the status notions in simple examples, expressing rules in words for clarity.
\begin{example}[Status] Consider trace $\rho=(L_{T_0},...,L_{T_f})$ produced by a sample collection robot. Suppose we select a suffix $\rho^{t_0...}$ to analyze, where $T_0 \leq t_0 \leq T_f$. Then we have the following status examples:
\par \textbf{Active}: Let $\varphi=\textup{``Headlight always on at night''}$ and suppose ${\textup{``night'', ``light on''}\in L_{t}}$ for all $t\geq t_0$. Then $\varphi$ is active at all $t\geq t_0$.
\par \textbf{Satisfied}: Let ${\varphi=\textup{``Eventually deposit sample''}}$. Suppose ${\textup{``sample deposited''}\in L_{t}}$,\\ ${\textup{``sample deposited''}\not\in L_{t'}}$ for all $t'<t$. Then $\varphi$ is satisfied at $t$.
\par \textbf{Inactive} (done): Let ${\varphi=\textup{``Eventually deposit sample''}}$ and suppose ${\textup{``sample deposited''}\in L_{t'}}$ for some $t'<t$. Then $\varphi$ is inactive at $t$, since $\varphi$ was already satisfied.
\par \textbf{Inactive} (not triggered): Let ${\varphi=\textup{``If raining, stop''}}$ and suppose ${\textup{``rain''} \not\in L_{t}}$. Then $\varphi$ is inactive at $t$, since precondition \textup{``rain''} is not occurring and stopping is not necessary.
\par \textbf{Violated}: Let ${\varphi=\textup{``Battery always above 10\%''}}$ and suppose ${\textup{``battery empty''}\in L_{t}}$. Then $\varphi$ is violated at all $t$.
\end{example}
\par Finally, we can express the set of all times in $\{t_0,...,T_f\}$ for which $\varphi$ has status $q$ on $\rho^{t_0...}$ as a \textit{timeset}
\begin{equation}\label{taudef}
   \tau^q= \{t \in \{t_0,...,T_f\}\ |\ t \text{ has status } q\}.
\end{equation}
If we recover such time sets for a rule $\varphi$, it is possible to check its status at any moment in an agent trajectory; moreover, time sets for each argument of $\varphi$ may help to track progress ``through'' the rule, such as completion of subtasks. 
\par \textbf{Formal Problem Statement.} For an LTL formula $\varphi$ and a plan trace $\rho^{T_0...T_f}$, an explainer must be able to provide:
\begin{itemize}
    \item \textbf{Status} of $\varphi$ at $t$ on $\rho^{t_0...}$ for all ${\{t_0, t\in\{T_0,...,T_f\}\ |\ t_0 \leq t \}}$
    \item \textbf{Status timesets} $\tau^a,\tau^s,\tau^i,\tau^v$ of $\varphi$ for each ${t_0\in\{T_0,...,T_f\}}$
    \item \textbf{Both of the above} for any argument $\varphi_j$ of $\varphi$.
\end{itemize}

In this paper, we propose an algorithm that returns this information on demand and begin to explore its applications.

\section{Rule Status Assessment}\label{aabyformula}
We wish to evaluate rule status for all arguments of a formula. To accomplish this in a structured manner, formulas are first decomposed into a tree structure and then passed through a modular algorithm.

\subsection{Modules and Algorithm}\label{algsection}

\begin{figure}[t]
\centering
\includegraphics[width=.4\textwidth]{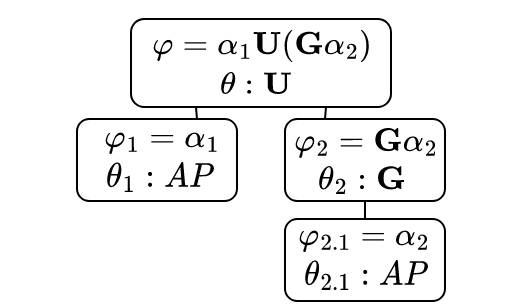}
\caption{Example formula tree for $\varphi=\alpha_1\mathbf{U}(\mathbf{G}\alpha_2)$.}
\label{treeEx}
\end{figure}

 LTL formulas $\varphi$ are commonly structured as trees with $\varphi$ as the base node \cite{lemieux2015general} \cite{kim2019bayesian}. Here, the children of a parent node are the arguments of the parent formula, until each branch terminates in a leaf containing only a label. We additionally store the type of the weakest operator in $\varphi_j$ for each node, denoted $\theta_j$; in the case where $\varphi_j = \alpha$, we assign type $\theta_j = AP$ (``atomic proposition'') to denote that the node contains a label. An example is given in Figure \ref{treeEx}. 

Each node of a tree contains a formula consisting of one or more arguments bound by an operator. The operator type determines how the argument truth values map to a status; for instance, consider a trace $\rho$ where a formula $\varphi_j$ is true at a single time step $t'$. Here, a node with $\varphi = \mathbf{G}\varphi_j$ (type $\mathbf{G}$) is violated at all $t$; a node $\varphi = \mathbf{F}\varphi_j$ (type $\mathbf{F}$) is active until satisfaction at $t'$. Thus, we establish separate modules for every type, each with inputs $\varphi$, $t_0$, $\rho^{t_0...}$ and outputs $\tau^a,\tau^s,\tau^i,\tau^v$ to satisfy Definition \ref{activedef}.
 \par Our proposed algorithm works node-by-node, beginning with all leaf nodes and working up each branch until $\tau$ has been calculated for the entire tree (child nodes are always instantiated before their parents). The algorithm then accepts queries and returns status at the desired time steps.
 \par The RSA algorithm is described informally below. A more rigorous definition is given in pseudocode in the Appendix.

\begin{tcolorbox}[colback=gray!10!white,colframe=gray!75!black,left=0pt,right=10pt,title=Rule Status Assessment Algorithm]
\begin{enumerate}
\item Construct tree for LTL rule $\varphi$.
\item Find $\tau^a,\tau^s,\tau^i,\tau^v$ of each node for all $\rho^{t_0...}$, ${t_0\in\{T_0,...,T_f\}}$:
\begin{enumerate}
\item For each distinct label $\alpha$ appearing in any leaf, pass $\rho^{t_0...}$, ${\varphi=\alpha}$ into the $AP$ module for all $t_0\in\{T_0,...,T_f\}$. Store outputs $\tau$ for all leaves containing this $\alpha$.
\item For leaf $\varphi_{...x.y}$, check parent $\varphi_{...x}$. If $\tau$ is already stored for all children of $\varphi_{...x}$, proceed to (c); otherwise, move to next leaf and repeat (b).
\item Identify $\theta$ of $\varphi_{...x}$. Pass $\varphi_{...x}$, $\rho^{t_0...}$ into $\theta$ module for all $t_0\in\{T_0,...,T_f\}$; store output $\tau$ for $\varphi_{...x}$.
\item Check parent of $\varphi_{...x}$. If it exists and all its children are instantiated, repeat step c) for this parent. Otherwise, move to next leaf and begin again from step (b).
\end{enumerate}
\item Answer status queries:
\begin{enumerate}
    \item Accept $t^*, t^*_0$ and a node (e.g., $\varphi_{x.y}$) as input.
    \item Return status $q$ if $t^*\in\tau^q$ for $\rho^{t_0^*...}$.
\end{enumerate}
\end{enumerate}
\end{tcolorbox}

\par The temporal and logical operator modules are defined in Tables \ref{tab:tempmodules} and \ref{tab:logmodules}, respectively. 
\par Finally, we claim that the modules as defined will correctly assign status values according to our status definitions. This is captured by the following theorem.

\begin{theorem}[Correctness of Modules]\label{correctness}
Consider a formula $\varphi_x$ with type $\theta$. Given trace segment $\rho^{t_0...}$ and all status timesets $\tau_{x.y}$ evaluated at $t_0$, where $\varphi_{x.y}$ are the children of $\varphi_x$, the module of type $\theta$ returns all timesets $\tau_x$ for $\varphi_x$ which satisfy Definition \ref{activedef}.
\end{theorem}

The full proof of Theorem \ref{correctness}, addressed module-by-module, is located in the Appendix.

\begin{table}[t]
\centering
\begin{tabular}{lp{12 cm}}
\toprule
\textbf{Not} $(neg)$& LTL: $\varphi=\neg\varphi_1$\\
\midrule
Initialize: & Load all $\tau_1^q$ of child node $\varphi_1$ on $\rho^{t_0...}$.\\
If $\tau^v_1=\emptyset:$&Set $\tau^v=\{t_0,...,T_f\}$; $\tau^a,\tau^s,\tau^i=\emptyset$.\\
Else:&Set $\tau^a,\tau^s=\{t_0\}, \tau^i=\{t_0,...,T_f\}\setminus \tau^a$. Set $\tau^v=\emptyset$.\\
\midrule
\textbf{Or} $(or)$&LTL: $\varphi=\varphi_1\vee\varphi_2$\\
\midrule
Initialize: & Load all $\tau_1^q,\tau_2^q$ of child nodes $\varphi_1, \varphi_2$ on $\rho^{t_0...}$.\\
$^*$If $\tau^v_1,\tau^v_2\neq\emptyset$:&Set $\tau^v=\{t_0,...,T_f\}$; $\tau^a,\tau^s,\tau^i=\emptyset$. \\
Else:&Set $\tau^a,\tau^s=\{t_0\}, \tau^i=\{t_0,...,T_f\}\setminus \tau^a$. Set $\tau^v=\emptyset$.\\
\midrule
\textbf{And}  $(and)$&LTL: $\varphi=\varphi_1\wedge\varphi_2$\\
\midrule
&Same as $\vee$ with $^*$ replaced by:\\ 
&If $\tau^v_1$ or $\tau^v_2\neq\emptyset$: (...)\\
\midrule
\textbf{Implication}  $(\rightarrow)$&LTL: $\varphi=\varphi_1\rightarrow\varphi_2$\\
\midrule
Initialize: & Load all $\tau_1^q$ of child node $\varphi_1$ on $\rho^{t_0...}$.\\
If $\tau^v_1\neq\emptyset$:&Set $\tau^a,\tau^s,\tau^v=\emptyset$; $\tau^i=\{t_0,...,T\}\setminus\tau^a.$ \textbf{Exit module.}\\
Else:&Load all $\tau_2^q$ of child node $\varphi_2$ on $\rho^{t_0...}$.\\
If $\tau^v_2\neq\emptyset$:&Set $\tau^v=\{t_0,...,T_f\}$; $\tau^a,\tau^s,\tau^i=\emptyset$.\\
Else:&Set $\tau^a, \tau^s=\{t_0\}; \tau^i=\{t_0,...,T_f\}\setminus\tau^a$. Set $\tau^v=\emptyset$.\\
\bottomrule\\
\end{tabular}
\caption{\label{tab:logmodules} Logical operator module definitions.}
\end{table}

\begin{table}[h!]
\centering
\begin{tabular}{lp{12 cm}}
\toprule
\textbf{Atomic}  $(AP)$ & LTL: $\varphi=\alpha$\\
 \midrule

If $\alpha\in L_{t_0}$:&Set $\tau^a$, $\tau^s=\{t_0\}$, $\tau^i = \{t_0,...,T_f\}\setminus \tau^a$. Set $\tau^v = \emptyset$.\\
Else: &Set $\tau^v=\{t_0,...,T_f\}$; $\tau^a,\tau^s,\tau^i= \emptyset$.\\
 \midrule
\textbf{Next}  $(X)$&LTL: $\varphi=\mathbf{X}\varphi_1$\\
\midrule
Initialize: & Load all $\tau_1^q$ of $\varphi_1$ on $\rho^{t_0+1...}$. \\
If $\tau_1^v = \emptyset$:& Set $\tau^a = \{t_0, t_0+1\}$; $\tau^s = \{t_0+1\}$, $\tau^i = \{t_0,...,T_f\}\setminus \tau^a$. Set $\tau^v = \emptyset$.\\
Else:& Set $\tau^v = \{t_0,...,T_f\}$; $\tau^a,\tau^s,\tau^i=\emptyset$.\\
\midrule
\textbf{Eventual} $(F)$&LTL: $\varphi=\mathbf{F}\varphi_1$\\
\midrule
For $t_0^1=t_0...T_f$:& Load all $\tau_1^q$ of $\varphi_1$ on $\rho^{t_0^1...}$. Continue until $t_0^1=T_f$ or $\tau^v_1 = \emptyset$.\\
If $\tau_1^v=\emptyset$:&Set $\tau^a = \{t_0,...,t_0^1\}$; $\tau^s = \{t_0^1\}$; $\tau^i=\{t_0,...,T_f\}\setminus\tau^a$. Set $\tau^v=\emptyset$. \textbf{Exit module.}\\
 Else:&Set $\tau^v = \{t_0,...,T_f\}$; $\tau^a,\tau^s,\tau^i=\emptyset$.\\
\midrule
\textbf{Global} $(G)$&LTL: $\varphi=\mathbf{G}\varphi_1$\\
\midrule
For $t_0^1=t_0...T_f$:& Load all $\tau^q_1$ of child node $\varphi_1$ on $\rho^{t_0^1...}$. \\
&Continue until $t_0^1=T_f$ or $\tau_1^v \neq \emptyset$.\\
If $\tau_1^v\neq\emptyset$:&Set $\tau^v = \{t_0,...,T_f\}$; $\tau^a,\tau^s,\tau^i=\emptyset$. \textbf{Exit module.}\\
 Else:&Set $\tau^a=\{t_0,...,T_f\}; \tau^s=\{T_f\};\tau^i,\tau^v=\emptyset$.\\
 \midrule
 
 \textbf{Until} $(U)$&LTL: $\varphi=\varphi_1\mathbf{U}\varphi_2$\\
 \midrule
For $t'=t_0...T_f$:& Load all $\tau_1^q, \tau_2^q$ of $\varphi_1$ and $\varphi_2$ on $\rho^{t'...}$. Continue until $\tau_2^v = \emptyset$ or $\tau^v_1\neq\emptyset$ or $t'=T_f$.\\
If $\tau_2^v = \emptyset$:&Set $\tau^a = \{t_0,...,t'\}; \tau^s=\{t'\}; \tau^i=\{t',...,T_f\}\setminus\tau^a$. Set $\tau^v=\emptyset$.\\
$^*$Else: &Set $\tau^v=\{t_0,...,T_f\}$; $\tau^a,\tau^s,\tau^i=\emptyset$.\\

 \midrule
 \textbf{W. until} $(W)$&LTL: $\varphi=\varphi_1\mathbf{W}\varphi_2$\\
 \midrule
 &Same as $\mathbf{U}$ with $^*$ replaced by:\\
&Else: Set $\tau^a=\{t_0,...,T_f\}; \tau^s=\{T_f\}, \tau^i = \emptyset$. Set $\tau^v=\emptyset$.\\
\midrule

 \textbf{S. release} $(M)$&LTL: $\varphi=\varphi_1\mathbf{M}\varphi_2$\\
 \midrule
 For $t'=t_0...T_f$:& Load all $\tau_1^q, \tau_2^q$ of $\varphi_1,\varphi_2$ on $\rho^{t'...}$. Continue until $\tau_1^v = \emptyset$ or $\tau^v_2\neq\emptyset$ or $t'=T_f$.\\
If $\tau_2^v,\tau_1^v = \emptyset$:&Set $\tau^a = \{t_0,...,t'\}; \tau^s=\{t'\}; \tau^i=\{t',...,T_f\}\setminus\tau^a$. Set $\tau^v=\emptyset$.\\
Else: &Set $\tau^v=\{t_0,...,T_f\}$; $\tau^a,\tau^s,\tau^i=\emptyset$.\\
\midrule
 \textbf{Release} $(R)$&LTL: $\varphi=\varphi_1\mathbf{R}\varphi_2$\\
 \midrule
 For $t'=t_0...T_f$:& Load all $\tau_1^q, \tau_2^q$ of $\varphi_1,\varphi_2$ on $\rho^{t'...}$. Continue until $\tau_1^v = \emptyset$ or $\tau^v_2\neq\emptyset$ or $t'=T_f$.\\
If $\tau_1^v,\tau_2^v = \emptyset$:&Set $\tau^a = \{t_0,...,t'\}; \tau^s=\{t'\}; \tau^i=\{t',...,T_f\}\setminus\tau^a$. Set $\tau^v=\emptyset$.\\
Else if $\tau^v_2\neq\emptyset$: &Set $\tau^v=\{t_0,...,T_f\}$; $\tau^a,\tau^s,\tau^i=\emptyset$.\\
Else: &Set $\tau^a=\{t_0,...,T_f\}; \tau^s=\{T_f\}, \tau^i = \emptyset$. Set $\tau^v=\emptyset$.\\
\bottomrule\\
\end{tabular}
\caption{\label{tab:tempmodules} Temporal operator module definitions.}
\end{table}

\subsection{Algorithm Complexity} \subsection{Complexity} The RSA algorithm determines status by checking the truth of rule nodes at individual time steps in the trace. The number of total evaluations depends on the length of the trace, the number of nodes and operator types contained in the LTL specification, and the data itself. An upper bound on the number of evaluations can nonetheless be established by analyzing the tree structure in conjunction with the modules. The result is Theorem \ref{thm:rsacomplexity}.
\begin{theorem}[Rule Status Assessment Complexity]\label{thm:rsacomplexity}
For a formula tree with depth L, base node at level $l=0$, and maximum 2 children per node evaluated on trace $\rho$, complexity has upper bound
    \begin{equation}\label{complexity}
    \mathcal{O}(2^{L}|\rho|^2).
\end{equation}
\end{theorem}

\begin{proof}
    For each node $\varphi_{...x}$ in a tree, the corresponding module is called once. This module produces $\tau$ for every $t_0$ in $\{T_0,...,T_f\}$, resulting in $\leq|\rho|$ iterations. At each $t_0$, the module performs up to $T_f-t_0\leq |\rho|$ evaluations per argument $\varphi_{...x.y}$. Thus, any node with a binary operator (two arguments) performs $n\leq 2|\rho|^2$ computations. Therefore, a tree with depth L, base node at level $l=0$, and maximum 2 children per node, the number of nodes $n_L$ at level $L$ will satisfy $n_L \leq 2^L$. A bound on complexity is thus of order $2^L(2|\rho|^2)$, or 
    \begin{equation*}
    \mathcal{O}(2^{L+1}|\rho|^2).
    \end{equation*}
    for the full tree. However, recall that the final level $l=L$ of the tree must consist of atomic nodes, which require only $|\rho|$ evaluations each. Assuming $2^L$ nodes in the final level, the complexity may thus be tightened to $\leq2^L|\rho| + 2|\rho|^22^{L-1}$, or
\begin{equation}
    \mathcal{O}(2^{L}|\rho|^2).
\end{equation}
\end{proof}

Note that logical operators with more than 2 arguments (e.g. $\varphi_1\vee\varphi_2\vee\varphi_3$) can be rewritten as a tree of their groupings (e.g., $\varphi_1\vee(\varphi_2\vee\varphi_3)$), with $L$ adjusted appropriately. Note as well that, as discussed in \cite{brindise2023pointwise}, the upper bound is not typically achieved, as evidenced by reasonable runtimes in experiment. 
\par One clear example is the conservative bound on $n_L$; if a tree of depth $L$ consists of only single-argument operators, the tree will have exactly $L+1$ nodes, requiring up to $L|\rho|^2+(L+1)|\rho|$ computations. This is less than the bound in \eqref{complexity} by a factor of $2^L$.

\section{Application}
In this section, we apply status assessment on hypothetical runs of two agents using the rule status assessment algorithm located at \nolinkurl{https://github.com/n-brindise/PiT-diagnostics}. We also suggest a basic heuristic to identify times $t^*,t^*_0$ of potential interest in a trace. \textbf{Important update:} for examples in simulation, we recommend referencing our recent publication in L4DC 2024 \cite{brindise2024pointwise}.

\subsection{Query Heuristics} Status assessment allows for a large number of queries $(\mathcal{O}(2^{l+1}|\rho|^2))$; thus, identification of informative queries from this large set is desirable. A first-order approach simply monitors the $\tau$ values as the algorithm iterates over $t_0\in\{T_0,...,T_f\}$ and stores any $t_0$ where $\rho$ changes status to \textit{active} or \textit{violated} from the preceding $t_0$. We denote the set of all $t^*_0$ of interest by $\tau^*$.

\subsection{Muddy Yard Example} 
\begin{figure}[t]
\centering
\includegraphics[width=.4\columnwidth]{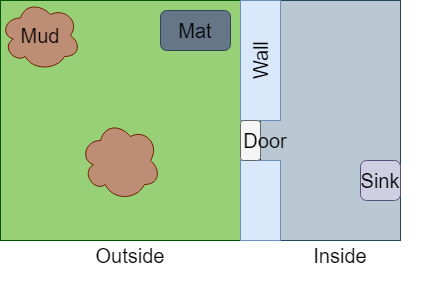}
\caption{``Muddy yard'' example domain, representing a yard (``outside'') with two muddy patches and a mat, as well as a home interior (``inside'') with a sink. Outside and Inside are separated by an untraversable wall.}
\label{muddyyard1}
\end{figure}

\begin{figure}[t]
\centering
\includegraphics[width=.4\columnwidth]{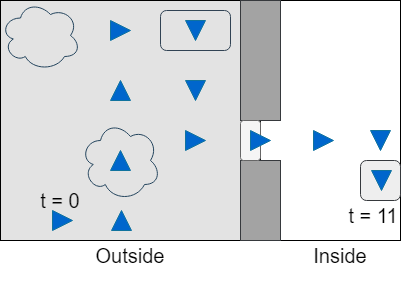}
\caption{Example path over the muddy yard domain. Each triangle represents the agent position and direction of travel at a single time step.}
\label{muddyyard2}
\end{figure}

The real world-inspired ``muddy yard'' scenario (Fig. \ref{muddyyard1}) features an ``outdoor'' area with two ``muddy'' sections and an outdoor ``mat,'' as well as an ``indoor'' area with a ``sink.'' The two areas are separated by a ``wall'' with a ``door.'' We formulate the domain as a Kripke structure with set $S$ of states and set $P$ of labels:
\begin{align*}
S = \{\sigma_1 &= \text{grass}\quad \sigma_2 =\text{mud}\quad \sigma_3=\text{mat}\quad \sigma_4 = \text{floor}\\
\quad \sigma_5 &= \text{sink}\quad \sigma_6 = \text{wall}\quad
\sigma_7 =\text{door} \}\\
P = \{\alpha_1 &= \text{outside}\quad \alpha_2 =\text{inside}\quad \alpha_3=\text{muddy}\\
\alpha_4 &= \text{wiped}\quad \alpha_5 = \text{washed}\quad \alpha_6 = \text{impassable}\}
\end{align*}
 where $I=S$. The transition relation $\mathcal{T}$ exclusively allows the agent to transition between adjacent states (e.g., $(\sigma_1,\sigma_2)\in \mathcal{T}$, but $(\sigma_1,\sigma_5)\not\in \mathcal{T}$). Finally, the labeling function $\mathcal{L}$ is
 \begin{align*}
\mathcal{L}(\sigma_1) &= \{\alpha_1\}\quad \mathcal{L}(\sigma_2) =\{\alpha_1,\alpha_3\}\quad \mathcal{L}(\sigma_3)=\{\alpha_1,\alpha_4\}\\
\mathcal{L}(\sigma_4) &= \{\alpha_2\}\quad \mathcal{L}(\sigma_5) = \{\alpha_2,\alpha_5\}\quad \mathcal{L}(\sigma_6) = \{\alpha_6\}\\
\mathcal{L}(\sigma_7) &=\{\alpha_1,\alpha_2\}.
\end{align*}
We wish to analyze the agent run in Figure \ref{muddyyard2} subject to the following LTL rules:
\begin{enumerate}
    \item $\varphi = \mathbf{F}\alpha_1$  ``Eventually be outside''
    \item $\varphi = \mathbf{F}\alpha_2 \rightarrow \mathbf{F}\alpha_5$  ``If eventually inside, eventually wash up''
    \item $\varphi = \mathbf{G}(\alpha_3\rightarrow(\neg\alpha_2\mathbf{W}\alpha_4))$  ``Always, if muddy, do not go inside until wiping off''
    \item $\varphi = \mathbf{G}(\neg\alpha_6)$  ``Never enter an impassable place''
\end{enumerate}

From Figure \ref{muddyyard2}, the system run is
\begin{equation*}
\kappa=\sigma_1,\sigma_1,\sigma_2,\sigma_1,\sigma_1,\sigma_3,\sigma_1,\sigma_1,\sigma_7,\sigma_4,\sigma_4,\sigma_5,
\end{equation*}
which produces the trace
\begin{align*}
    \rho = &\{\alpha_1\}_0, \{\alpha_1\}_1,\{\alpha_1,\alpha_3\}_2, \{\alpha_1\}_3, \{\alpha_1\}_4, \{\alpha_1,\alpha_4\}_5,\\
    &\{\alpha_1\}_6, \{\alpha_1\}_7,\{\alpha_1,\alpha_2\}_8, \{\alpha_2\}_9,\{\alpha_2\}_{10}, \{\alpha_2,\alpha_5\}_{11}.
\end{align*}
We now use status assessment to answer a series of questions. (When listing timeset outputs, we omit any $\tau=\emptyset$ for clarity.)

\par \textbf{Question:} ``Which rules are active at time $t^*$?'' 
\par For $t_0=T_0$, Rules 1-4 have (nonempty) timesets
\begin{enumerate}
    \item $\tau^a = \{0\},\tau^s=\{0\},\tau^i=\{1,...,11\}$
    \item $\tau^a = \{0\},\tau^s=\{0\},\tau^i=\{1,...,11\}$
    \item $\tau^a=\{0,...,11\}, \tau^s = \{11\}$
    \item $\tau^a=\{0,...,11\}, \tau^s = \{11\}$
\end{enumerate}
At $t^*=3$, output is:
\begin{itemize}
    \item \textit{``Rules 1 and 2 are inactive''}
    \item \textit{``Rules 3 and 4 are active''}
\end{itemize} 
Intuitively, Rule 1 requires that the agent goes outside; this occurred already at $t=0$, hence the inactivity by $t=3$. Rule 2 requires that, if the agent will eventually go inside, it must also eventually wash up; both of these conditions are already true on $\rho^{0...}$, and thus inactive for $t^*>0$. This output is less intuitive, as it is not clear \textit{when} the agent will go inside or wash up.
 Finally, Rules 3 and 4 are global and must be active at every time step. For 2, 3, and 4, elaboration is likely necessary, prompting another question.
\par \textbf{Question: } ``Which arguments of rules are active at $t^*$?''
\par We first query Rule 3 at node $\varphi_{1}$. From heuristics, ${\tau^*=\{2\}}$, so we query $\varphi_{1}$ at ${t^*,t^*_0=2}$, which yields
\begin{itemize}
    \item $\tau^a_1 = \{2\},\tau^s_1=\{2\},\tau^i_1=\{0,1,3,...,11\}$
    \item \textit{``Rule 3.1 is active and satisfied (at $t^*=2$)''}
\end{itemize} 
This suggests that $\alpha_3 \rightarrow (\neg\alpha_2\mathbf{W}\alpha_4)$ is active on $\rho^{2...}$. Indeed, $\alpha_3$ (``muddy'') occurs at $t=2$, triggering the implication. Querying $\varphi_{1.2}$ at ${t_0^*=2, t^*=5}$, 
\begin{itemize}
    \item ${\tau^a_{1.2} = \{2,...,5\}},{\tau^s_{1.2}=\{5\}},{\tau^i_{1.2}=\{6,...,11\}}$
    \item \textit{``Rule 3.1.2 is active and satisfied (at $t^*=5$)''}
\end{itemize}
Intuitively, though Rule 3 always holds, it only takes direct effect starting at 2 (when the agent becomes muddy) and releases this effect at 5 (when the agent wipes itself off).
\par Moving on to Rule 2, we perform queries for $\varphi_1=\mathbf{F}\alpha_2$ and $\varphi_2=\mathbf{F}\alpha_5$, and we find that both are \textit{active} at $t_0^*=0,t^*=3$. This makes sense, as the agent has neither gone inside nor washed up by step $3$. At $t^*_0, t^*=9$, $\varphi_1$ is now \textit{inactive}, since the agent has been inside for several steps, but $\varphi_2$ remains \textit{active}, since it has yet to wash up. Finally, the agent washes up at $11$, and $t^*=11$ gives $\varphi_2$ \textit{satisfied}.

\subsection{Autonomous Vehicle Example} Autonomous vehicles must comply to safety guidelines, operational limits, and traffic laws, all of which are typically expressible via temporal logics \cite{9827153}. This example considers a hypothetical setting in which an autonomous vehicle completes three distinct trips, producing 3 traces over a large vocabulary ($|\mathcal{P}|=54$) and 90 time steps. 21 LTL formulas describe various rules for safety, legal, and trip requirements, such as ``reduce speed in construction zones'' and ``activate wipers if rain hazard present.'' The traces and list of rules are available in the Appendix.
\par Notably, the raw traces are visually difficult to parse. Single time steps contain lengthy sets of labels, for example \{leftmost, doors-closed, driver-awake, clear-ahead, want-turn-left, clear-right, lane-to-right, gas-low, intersection-ahead, green-ahead, reduced-speed, safe-stop\} at $\rho_1,t=36$. 

\begin{figure}[t]
\centering
\includegraphics[width=.5\columnwidth]{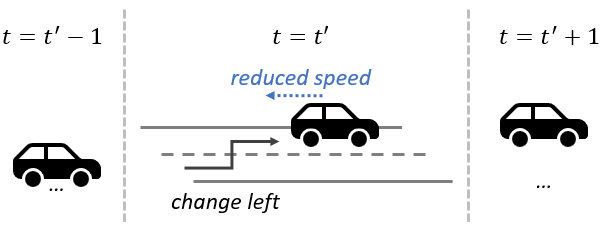}
\caption{Vehicle observed changing lanes and reducing speed at $t=t'$.}
\label{lanechange}
\end{figure}

\par To motivate our example, suppose we observe the vehicle \textbf{changes lanes to the left} and \textbf{reduces speed} at time $t'$ in all three trips, as shown in Figure \ref{lanechange}. We wish to investigate what has occurred in each trace that may explain this behavior.
\par As the traces are dense and the list of rules is long, we use status assessment. We place queries at $t_0^*,t^*=t'$ for the first argument of all 20 global rules $\mathbf{G}(...)$. 
\par For $\rho_1$, we find (1) \textbf{satisfied} $\text{gas-low}\rightarrow\neg(\text{goal})\mathbf{U}\text{gas-station}$, (2) \textbf{satisfied} $\text{want-left-turn}\wedge\text{near-intersection}\wedge\text{clear-left}\wedge\text{lane-to-left}\rightarrow \text{change-left}$ and (3) \textbf{satisfied} $\text{want-turn}\wedge\text{near-intersection}\rightarrow \text{reduced-speed}$. In words, the vehicle is heading to a gas station, approaching an intersection, and planning to turn left.
\par For $\rho_2$, we find (1) \textbf{satisfied} $\text{merge-left}\wedge\text{clear-left}\wedge\text{lane-to-left} \rightarrow \text{change-left}$ and (2) \textbf{satisfied} $\text{construction-zone} \rightarrow \text{reduced-speed}\mathbf{U}\text{leave-zone}$. Here, the vehicle has entered a construction zone and must merge.
\par Finally, for $\rho_3$, we have (1) \textbf{satisfied} $\text{short-follow-distance} \rightarrow \text{reduced-speed}$. We place an additional query at $t^*_0, t^*=t'-1$, which returns (2) \textbf{satisfied} $\text{short-follow-distance} \wedge(\text{clear-left}\wedge\text{lane-to-left})\rightarrow \mathbf{X}\text{change-left}$.
Thus, the vehicle encountered another vehicle which had dangerously reduced the following distance.
\par In all, status assessment processed a large set of rules and dense traces to produce brief diagnostic statements which highlighted the specific circumstances of each left-lane change, a promising aid to understand the situation.

\section{Discussion and Conclusion}\label{conclusion}
\par In this work, we introduce rule status assessment (RSA), a method for explanatory diagnostics for generic autonomous agent trajectories. RSA is applicable to any system expressible via Kripke structure, providing a novel framework to classify LTL rules of interest as active, satisfied, inactive, or violated at individual trajectory time steps. This makes it possible to track an agent's progress through subtasks and more; it also provides insight into the applicability of individual rules and reduces the total number of rules a human must consider at once.
\par Multiple directions exist for future work. Firstly, the broader application of RSA is a critical next step to demonstrate effectiveness for practical purposes, particularly in common settings such as Machine Learning (ML). Further application also calls for additional query heuristics; it is a challenge to identify useful queries to achieve insightful RSA diagnostics. Finally, the existing framework is entirely post-hoc; for certain families of systems, such as Reinforcement Learning agents, use of agent internals (i.e., Q function) could incorporate agent intentionality.

\bibliography{References}

\section{Appendix}
\subsection{Rule Status Assessment Algorithm}\label{appendixAlg}
\begin{algorithm}[H]
\SetAlgoLined
\SetKwInOut{Input}{Input}
\SetKwInOut{Output}{Output}
\SetKwInOut{print}{print}
\SetKwInOut{Return}{Return}
\SetKwFunction{makeTree}{makeTree}
\SetKwFunction{getLabels}{getLabels}
\SetKwFunction{getLeaves}{getLeaves}
\SetKwFunction{APmodule}{APmodule}
\SetKwFunction{populate}{populate}

\nl\Input{Trace $\rho$, rule $\varphi$}
\nl\Output{Status timesets $\tau$ for all nodes of $\varphi$}
\nl\KwData{Operator modules (Tables \ref{tab:logmodules} and \ref{tab:tempmodules})}
\nl\textbf{Initialize} $T_0\leftarrow0$; $T_F\leftarrow |\rho|$; ruleNode object class with the following properties:\\
\quad $.\theta \leftarrow ()$; $.\tau^a$\_list $\leftarrow ()$; ... $.\tau^v$\_list $\leftarrow$ ()\\
\quad .children $\leftarrow ()$; .parent $\leftarrow ()$; .formula $\leftarrow ()$; .populated $\leftarrow$ \texttt{False} \\
\nl ruleTree $\leftarrow$ \makeTree($\varphi$)  \quad \tcp{Decompose $\varphi$ into tree of RuleNode objects. Assign $\theta$, formula, children, and parent properties.}
\nl labelList $\leftarrow$ \getLabels($\varphi$);\quad  \tcp{Get list of all labels appearing in $\varphi$}

\nl labelNodes = dict()\quad \tcp{Dictionary of dummy ruleNode objects}

\nl\For{label in labelList}{
\nl    dummyNode $\leftarrow$ \textbf{new} ruleNode with $\theta \leftarrow AP$ \\
\nl    \For{$t_0$ in ($T_0,...,T_f$)}{
\nl    $\rho^{t_0...}\leftarrow \rho[t_0:T_F]$ \tcp{Segment of trace from $t_0$ to end}
\nl    $\tau^a,\tau^s,\tau^i,\tau^v$ $\leftarrow$ \APmodule($\rho^{t_0...},$ label)\\
\nl    \textbf{Append} $\tau_a$ to dummyNode.$\tau^a$\_list; ... \tcp{do for all 4 $\tau$}
    }
\nl    labelNodes[label] $\leftarrow$ dummyNode
}
\nl done $\leftarrow$ \texttt{False} \tcp{Now work up the rule tree to fill in status information.}
\nl leafList $\leftarrow$ \getLeaves(ruleTree)\quad \tcp{get list of all ruleNodes with $\theta=AP$ from ruleTree}

\nl currentNode $\leftarrow$ leafList[0]; leafCounter $\leftarrow 0$\\
\nl\While{not done}{
\nl    \uIf{currentNode.children empty}{
\nl        label $\leftarrow$ currentNode.formula     \tcp{Must be a leaf node}
\nl        currentNode.$\tau^a$\_list $\leftarrow$ labelNodes[label].$\tau^a$\_list;... \tcp{do for all 4 $\tau$}
\nl        currentNode.populated $\leftarrow$ \texttt{True}\\
\nl        \textbf{Store} currentNode in ruleTree\\
\nl        leafCounter$++$; currentNode $\leftarrow$ leafList[leafCounter] \tcp{move to next leaf}
    }
\nl    \uElse{
\nl        children $\leftarrow$ currentNode.children \tcp{Not a leaf node}
\nl        \uIf{child.populated for all child in children}{\tcp{all child nodes are populated already}
\nl            \populate(currentNode, $\rho$); \quad\textbf{Store} currentNode in ruleTree\\
\nl            currentNode $\leftarrow$ currentNode.parent
        }
\nl        \uElse{
\nl        leafCounter$++$; currentNode $\leftarrow$ leafList[leafCounter] \tcp{move to next leaf}}
    }   
        
\nl    \uIf{currentNode.parent empty}{
\nl        done $\leftarrow$ \texttt{True}
    }
}
\nl\Return{ruleTree}
\caption{Rule Status Assessment.}\label{RSAalgorithm}
\end{algorithm}

\begin{algorithm}
\nl\textbf{function} \texttt{populate}(node, $\rho$):\\
\nl    nodeType $\leftarrow$ currentNode.$\theta$\\
\nl    \For{$t_0$ in ($T_0,...,T_f$)}{
\nl        $\rho^{t_0...}\leftarrow \rho[t_0:T_F]$\\
\nl        $\tau^a,\tau^s,\tau^i,\tau^v$ $\leftarrow$ \texttt{toModule}($\rho^{t_0...},$ $\theta$, node.children) \tcp{Get output from appropriate module}
\nl        \textbf{Append} $\tau_a$ to node.$\tau^a$\_list; ... \tcp{do for all 4 $\tau$}
    }
\nl    node.populated $\leftarrow$ \texttt{True}
\caption{RSA \texttt{populate} Function.}\label{RSAalgorithmPopulate}
\end{algorithm}
\FloatBarrier

\subsection{Algorithm Soundness by Module}\label{appendix}

 \textbf{Theorem \ref{correctness}}: \textit{Consider a formula $\varphi_x$ with type $\theta$. Given trace segment $\rho^{t_0...}$ and all status timesets $\tau_{x.y}$ evaluated at $t_0$, where $\varphi_{x.y}$ are the children of $\varphi_x$, the module of type $\theta$ returns all timesets $\tau_x$ for $\varphi_x$ which satisfy Definition \ref{activedef}.}

Each module is considered separately. For each, Definition \ref{FiniteTraceDef} is extended as necessary to identify the specific conditions for $\rho^{t'...}\models\varphi$ given a particular operator. It is then possible to evaluate the module's assignment of status against the criteria set by Definition \ref{activedef}. 

\par By the proposed algorithm, all modules take the following inputs:
\begin{itemize}
    \item \textbf{Initial and final times} $t_0, T_f$, respectively
    \item \textbf{Trace} $\rho^{t_0...} = (L_{t_0},...,L_{T_f})$
    \item \textbf{Status timesets} $\tau^q_x$ for all status types $q$ for each argument $\varphi_x$ of $\varphi$
\end{itemize}
Since the status assessment algorithm works up from child to parent nodes, status timesets for the arguments of a formula will always be instantiated by the time the formula module is called.

\par The relevant definitions for arbitrariness and preconditions are reproduced here as well for convenience.

\begin{tcolorbox}[colback=gray!10!white,colframe=white!80!black,title=\color{black}{Definition \ref{arbitraryDef}: Arbitrary Suffix}]
For LTL formula $\varphi$, its argument(s) $\varphi_j$, trace $\rho^{T_0...}$, and associated segment $\rho^{t_0...}$ where $T_0\leq t_0 \leq T_f$, we say that $\rho^{t...}$ is \textup{arbitrary} with respect to $\varphi$ if $\rho^{t_0...}\models\varphi$ is independent of the truth of $\rho^{t...}\models\varphi_j$ for all $t$ in $t'\leq t\leq T_f$ for some $t_0\leq t' \leq T_f$. 
\end{tcolorbox}

\begin{tcolorbox}[colback=gray!10!white,colframe=white!80!black,title=\color{black}{Definition \ref{preconditionDef}: Precondition}]
The precondition of an LTL formula $\varphi$ is defined as (1) $\varphi_1$ if $\varphi$ has the form $\varphi_1\rightarrow\varphi_2$ and (2) $\top$ otherwise.
\end{tcolorbox}

\begin{tcolorbox}[colback=gray!10!white,colframe=white!80!black,title=\color{black}{Definition \ref{activedef}: Rule Status}]
We define the following statuses for an LTL formula $\varphi$ given a $\rho^{T_0...}$ and $t_0,t\in\{T_0,...,T_f\}$, $t_0\leq t$:
\par $\varphi$ is \textbf{active} (a) at $t$ iff 
\par \textbf{(1a)} $\rho^{t_0...}\models\varphi$, 
\par \textbf{(2a)} $\rho^{t_0...}\models\psi$ for precondition $\psi$ of $\varphi$, and
\par \textbf{(3a)} $\rho^{t...}$ is not arbitrary. 

\par $\varphi$ is \textbf{satisfied} (s) at $t$ iff \par \textbf{(1s)} $\varphi$ \textup{active} at $t$ and 
\par \textbf{(2s)} $t=T_f$ or $\varphi$ not active at $t+1$.
\par $\varphi$ is \textbf{inactive} 
 (i) at $t$ iff 
 \par \textbf{(1i)} $\rho^{t_0...}\models\varphi$ and \par\textbf{(2i)} $\varphi$ is not \textup{active} at $t$. 
\par $\varphi$ is \textbf{violated} (v) at $t$ iff \par \textbf{(1v)} $\rho^{t_0...}\not\models\varphi$.
\end{tcolorbox}

\par We begin with the module for atomic propositions and logical operator modules. We then proceed to the temporal operator modules.

\begin{tcolorbox}[colback=gray!10!white,colframe=white!80!black,title=\color{black}{Definition \ref{FiniteTraceDef} and Extensions}]
    LTL formula $\varphi$ is true on finite trace $\rho^{t_0...} = (L_{t_0},...,L_{T_f})$, denoted $\rho^{t_0...}\models\varphi$, \textbf{if} $T_0\leq t_0 \leq T_f$ \textbf{and}:
        \par$\bullet$ $\rho^{t_0...} \models \alpha$ where $\alpha\in P\quad$ iff $\quad\alpha\in L_0$
        \par$\bullet$ $\rho^{t_0...}\models \neg \varphi\quad$ iff $\quad\rho^{t_0...}\not\models\varphi$
        \par$\bullet$ $\rho^{t_0...} \models \varphi_1 \wedge \varphi_2\quad$ iff $\quad\rho^{t_0...}\models\varphi_1$ and $\rho^{t_0...}\models\varphi_2$
        \par and for the temporal operators \textup{Next} $\mathbf{X}$ and \textup{Until} $\mathbf{U}$, 
    \par$\bullet$ $\rho^{t_0...} \models \mathbf{X}\varphi\quad$ iff $\quad \rho^{t_0+1...}\models\varphi$ \textbf{and} $t_0<T_f$
    \par$\bullet$ $\rho^{t_0...}\models\varphi_1\mathbf{U}\varphi_2\quad$ iff $\quad \exists i\geq t_0$ s.t. $\rho^{i...}\models\varphi_2$ and $\rho^{k...} \models \varphi_1$ for all $k$ s.t. $t_0\leq k < i$ \textbf{and} $i\leq T_f$

Extending definitions for the additional operators yields

 \par $\bullet$ $\rho^{t_0...} \models \varphi_1\vee\varphi_2 \quad$ iff $\quad \rho^{t_0...}\models\varphi_1$ \textbf{or} $\rho^{t_0...}\models\varphi_2$

  \par $\bullet$ $\rho^{t_0...} \models \varphi_1\rightarrow\varphi_2 \quad$ iff $\quad\rho^{t_0...}\not\models\varphi_1$ \textbf{or} $\rho^{t_0...}\models\varphi_2$

  \par $\bullet$ $\rho^{t_0...} \models \mathbf{F}\varphi \quad$ iff $\quad \exists i\geq t_0$ s.t. $\rho^{i...}\models\varphi$ for $i\leq T_f$

 \par  $\bullet$ $\rho^{t_0...} \models \mathbf{G}\varphi \quad$ iff $\quad \rho^{i...}\models\varphi$ for all $i$ where $t_0\leq i\leq T_f$

 \par  $\bullet$ $\rho^{t_0...} \models \varphi_1\mathbf{W}\varphi_2 \quad$ iff $\quad (\exists i\geq t_0$ s.t. $\rho^{i...}\models\varphi_2$ and $\rho^{k...} \models \varphi_1$ for all $t_0\leq k < i$ \textbf{and} $i\leq T_f)$ \textbf{or} $(\rho^{i...}\models\varphi_1$ for all $i$ where $t_0\leq i\leq T_f)$
 
  \par $\bullet$ $\rho^{t_0...}\models\varphi_1 \mathbf{R} \varphi_2\quad$ iff $\quad$ for all $i$ s.t. $t_0\leq i\leq T_f$, $(\rho^{i...}\models\varphi_2)\vee (\exists k<i$ s.t. $\rho^{k...}\models\varphi_1)$

  \par $\bullet$ $\rho^{t_0...}\models\varphi_1 \mathbf{M} \varphi_2\quad$ iff $\quad\exists i,t_0\leq i\leq T_f$ s.t. $(\rho^{i...}\models \varphi_1) \wedge (\rho^{i...}\models\varphi_2) \wedge (\rho^{k...} \models \varphi_2$ for all $k$ s.t. $t_0\leq k < i)$
\end{tcolorbox}

  \section{Logical modules}

\subsection{Atomic proposition}

\begin{table}[h]
\centering
\begin{tabular}{lp{13.0 cm}}
\toprule
\textbf{Atomic}  $(AP)$ & LTL: $\varphi=\alpha$\\
 \midrule
If $\alpha\in L_{t_0}$:&Set $\tau^a$, $\tau^s=\{t_0\}$, $\tau^i = \{t_0,...,T_f\}\setminus \tau^a$. Set $\tau^v = \emptyset$.\\
Else: &Set $\tau^v=\{t_0,...,T_f\}$; $\tau^a,\tau^s,\tau^i= \emptyset$.\\

\bottomrule\\
\end{tabular}
\caption{\label{tab:apmodule} Atomic proposition module definition.}
\end{table}
\FloatBarrier

\begin{tcolorbox}[colback=gray!10!white,colframe=white!80!black,title=\color{black}{Definition \ref{FiniteTraceDef}: Formula Evaluation on Finite Traces (AP)}]
 LTL formula $\varphi$ is true on finite trace $\rho^{t_0...} = (L_{t_0},...,L_{T_f})$, denoted $\rho^{t_0...}\models\varphi$, if $T_0\leq t_0 \leq T_f$ \textbf{and}:
        \par$\rho^{t_0...} \models \alpha$ where $\alpha\in P\quad$ \textbf{iff} $\quad\alpha\in L_0$
\end{tcolorbox}

\begin{tcolorbox}[colback=gray!50!white,colframe=white!60!black,title=Properties of $\varphi$]
\par \textbf{Precondition}: $true$ (Def \ref{preconditionDef})
\par \textbf{Arbitrary suffixes}: all $\rho^{t...}$ such that $t_0<t\leq T_f$ (Def \ref{arbitraryDef})
\begin{itemize}
\item Consider $t=t_0$. Clearly the truth of $\rho^{t_0...}\models\alpha$ depends on the truth of itself, so $\rho^{t_0...}$ is \textbf{not} arbitrary. However, since $\rho^{t_0...}\models\alpha$ depends solely on $L_{t_0}$, all other $L_t$ with $t>t_0$ do not affect the truth of $\rho^{t_0...}\models\alpha$. Thus, all $\rho^{t...}$ with $t>t_0$ are arbitrary.
\end{itemize}
\end{tcolorbox}

\begin{proof}[Proof: Soundness of AP Module] We examine the outputs of the AP module status by status. The status definition conditions are labeled \textbf{(1a)}, \textbf{(2a)}, etc. as in the reproduction of Definition \ref{activedef} above.

\begin{table}[h]
\centering
\begin{tabular}{lp{13.0 cm}}
\multicolumn{2}{l}{\textbf{AP Active status timeset $\tau^a$}}\\
\toprule
\textbf{If} $\alpha\in L_{t_0}$:&\textbf{Set} $\tau^a=\{t_0\}$\\
\midrule
& \par $\bullet$ We have that $\rho^{t_0...}\models \varphi$ by Def. \ref{FiniteTraceDef} \textbf{(1a)}. 
\par $\bullet$ $\rho^{t_0...}\models true$ \textbf{(2a)}. 
\par $\bullet$ All $\rho^{t...}$ are arbitrary except at $t=t_0$ for $t_0\leq t \leq T_f$ \textbf{(3a for $t_0$)}.
\par Therefore, \textbf{only $t=t_0$} meets all conditions for active status, and $\tau^a=\{t_0\}$. \qed   \\
\midrule
\textbf{Else}:&\textbf{Set} $\tau^a= \emptyset$\\
\midrule
& $\bullet$ If $\alpha\not\in L_{t_0}$, then $\rho^{t_0...}\not\models\alpha$. 
\par Thus, (1a) is not met for any $t$, and $\tau^a=\emptyset$. \qed  \\

\bottomrule\\
\end{tabular}
\end{table}
\FloatBarrier

\begin{table}[h]
\centering
\begin{tabular}{lp{13.0 cm}}
\multicolumn{2}{l}{\textbf{AP Satisfied status timeset $\tau^s$}}\\
\toprule
\textbf{If} $\alpha\in L_{t_0}$:&\textbf{Set} $\tau^s=\{t_0\}$\\
\midrule
& $\bullet$ $\varphi$ is active at $t_0$ \textbf{(1s for $t_0$)}. 
\par $\bullet$ $\varphi$ not active at $t>t_0$ \textbf{(2s for $t_0$)}. For all $t>t_0$, (1s) is not met. 
\par Therefore, \textbf{only} $t=t_0$ meets all conditions for satisfied status, and $\tau^s=\{t_0\}$. \qed  \\
\midrule
\textbf{Else}:&\textbf{Set} $\tau^s= \emptyset$\\
\midrule
& Since $\varphi$ is not active at any $t$, (1s) is not met for any $t$, so $\tau^s=\emptyset$. \qed  \\

\bottomrule\\
\end{tabular}
\end{table}
\FloatBarrier

\begin{table}[h]
\centering
\begin{tabular}{lp{13.0 cm}}
\multicolumn{2}{l}{\textbf{AP Inactive status timeset $\tau^i$}}\\
\toprule
\textbf{If} $\alpha\in L_{t_0}$:&\textbf{Set} $\tau^i=\{t_0,...,T_f\}\setminus t_0$\\
\midrule
& $\bullet$ $\rho^{t_0...}\models\alpha$ by Def. \ref{FiniteTraceDef} \textbf{(1i)}.
\par Therefore, $\tau^i =\{t_0,...,T_f\}\setminus \tau^a$ (see \textbf{2i}). \qed  \\
\midrule
\textbf{Else}:&\textbf{Set} $\tau^i= \emptyset$\\
\midrule
& $\rho^{t_0...}\not\models\alpha$ by Def. \ref{FiniteTraceDef}, so (1i) is not met. Thus $\tau^i=\emptyset$. \qed  \\

\bottomrule\\
\end{tabular}
\end{table}
\FloatBarrier

\begin{table}[h]
\centering
\begin{tabular}{lp{13.0 cm}}
\multicolumn{2}{l}{\textbf{AP Violated status timeset $\tau^v$}}\\
\toprule
\textbf{If} $\alpha\in L_{t_0}$:&\textbf{Set} $\tau^v=\emptyset$\\
\midrule
&$\rho^{t_0...}\models\alpha$ by Def. \ref{FiniteTraceDef}, so \textbf{no} $t$ meets (1v) for violated status and $\tau^v=\emptyset$. \qed  \\
\midrule
\textbf{Else}:&\textbf{Set} $\tau^v= \{t_0,...,T_f\}$\\
\midrule
& $\rho^{t_0...}\not\models\alpha$ by Def. \ref{FiniteTraceDef} \textbf{(1v)}. Thus $\tau^v=\{t_0,...,T_f\}$. \qed  \\

\bottomrule\\
\end{tabular}
\end{table}
\FloatBarrier

\end{proof}


\subsection{Not}

\begin{table}[h]
\centering
\begin{tabular}{lp{13.0 cm}}
\toprule
\textbf{Not} $(neg)$& LTL: $\varphi=\neg\varphi_1$\\
\midrule
Initialize: & Load all $\tau_1^q$ of child node $\varphi_1$ on $\rho^{t_0...}$.\\
If $\tau^v_1=\emptyset:$&Set $\tau^v=\{t_0,...,T_f\}$; $\tau^a,\tau^s,\tau^i=\emptyset$.\\
Else:&Set $\tau^a,\tau^s=\{t_0\}, \tau^i=\{t_0,...,T_f\}\setminus \tau^a$. Set $\tau^v=\emptyset$.\\

\bottomrule\\
\end{tabular}
\caption{\label{tab:negmodule} Not module definition.}
\end{table}
\FloatBarrier

\begin{tcolorbox}[colback=gray!10!white,colframe=white!80!black,title=\color{black}{Definition \ref{FiniteTraceDef}: Formula Evaluation on Finite Traces (neg)}]
    LTL formula $\varphi$ is true on finite trace $\rho^{t_0...} = (L_{t_0},...,L_{T_f})$, denoted $\rho^{t_0...}\models\varphi$, if $T_0\leq t_0 \leq T_f$ \textbf{and}:
        \par $\rho^{t_0...}\models \neg \varphi\quad$ \textbf{iff} $\quad\rho^{t_0...}\not\models\varphi$
\end{tcolorbox}

\begin{tcolorbox}[colback=gray!50!white,colframe=white!60!black,title=Properties of $\varphi$]
\par \textbf{Precondition}: $true$ (Def \ref{preconditionDef})
\par \textbf{Arbitrary suffixes}: all $\rho^{t...}$ such that $t_0<t\leq T_f$ (Def \ref{arbitraryDef})
\begin{itemize}
\item Consider $\varphi = \neg \varphi_j$. By Def. \ref{FiniteTraceDef}, $\rho^{t_0...}\models\varphi$ iff $\rho^{t_0...}\not\models\varphi_j$. Thus, $\rho^{t_0...}\models\varphi$ does not depend on whether $\rho^{t...}\models\varphi_j$ for any $t_0<t\leq T_f$, making all $\rho^{t...}$ with $t>t_0$ arbitrary.
\end{itemize}
\end{tcolorbox}

\begin{proof}[Proof: Soundness of Not Module] We examine the outputs of the Not module status by status. The status definition conditions are labeled \textbf{(1a)}, \textbf{(2a)}, etc. as in the reproduction of Definition \ref{activedef} above.

\begin{table}[h]
\centering
\begin{tabular}{lp{13.0 cm}}
\multicolumn{2}{l}{\textbf{Not Active status timeset $\tau^a$}}\\
\toprule
\textbf{If} $\tau^v_1=\emptyset$:&\textbf{Set $\tau^a=\emptyset$}\\
\midrule
& \par $\bullet$ $\tau^v_1=\emptyset$ on $\rho^{t_0...}$ iff $\rho^{t_0...}\models\varphi_1$ by Def. \ref{activedef}. Thus $\rho^{t_0...}\not\models\neg\varphi_1$.
\par Therefore, $\rho^{t_0...}\not\models\varphi$ by Def. \ref{FiniteTraceDef} and (1a) is not met for any $t$, so $\tau^a=\emptyset$.\qed  \\
\midrule
\textbf{Else}:&\textbf{Set} $\tau^a= \{t_0\}$\\
\midrule
& $\bullet$ $\rho^{t_0...}\not\models\varphi_1$, implying that $\rho^{t_0...}\models\varphi$ \textbf{(1a)}.
\par $\bullet$ $\rho^{t_0...}\models true$ \textbf{(2a)}.
\par $\bullet$ All $\rho^{t...}$ are arbitrary except at $t=t_0$ for $t_0\leq t \leq T_f$ \textbf{(3a for $t_0$)}.
\par Therefore, \textbf{only $t=t_0$} meets all conditions for active status, and $\tau^a=\{t_0\}$. \qed  \\

\bottomrule\\
\end{tabular}
\end{table}
\FloatBarrier

\begin{table}[h]
\centering
\begin{tabular}{lp{13.0 cm}}
\multicolumn{2}{l}{\textbf{Not Satisfied status timeset $\tau^s$}}\\
\toprule
\textbf{If} $\tau^v_1=\emptyset$:&\textbf{Set $\tau^s=\emptyset$}\\
\midrule
& \par Since $\varphi$ is not active at any $t$, (1s) is not met for any $t$, so $\tau^s=\emptyset$. \qed  \\
\midrule
\textbf{Else}:&\textbf{Set} $\tau^v= \emptyset$\\
\midrule
& $\bullet$ $\varphi$ is active at $t_0$ \textbf{(1s for $t_0$)}.
\par $\bullet$ $\varphi$ not active at $t>t_0$ \textbf{(2s for $t_0$)}. For all $t>t_0$, (1s) is not met. 
\par Therefore, \textbf{only} $t=t_0$ meets all conditions for satisfied status, and $\tau^s=\{t_0\}$. \qed  \\

\bottomrule\\
\end{tabular}
\end{table}
\FloatBarrier

\begin{table}[h]
\centering
\begin{tabular}{lp{13.0 cm}}
\multicolumn{2}{l}{\textbf{Not Inactive status timeset $\tau^i$}}\\
\toprule
\textbf{If} $\tau^v_1=\emptyset$:&\textbf{Set $\tau^i=\emptyset$}\\
\midrule
& Again, $\rho^{t_0...}\models\varphi_1$, so $\rho^{t_0...}\not\models\varphi$ by Def. \ref{FiniteTraceDef} and (1i) is not met. Thus $\tau^i=\emptyset$. \qed  \\
\midrule
\textbf{Else}:&\textbf{Set} $\tau^i= \{t_0,...,T_f\}\setminus \tau^a$\\
\midrule
& $\bullet$ $\rho^{t_0...}\not\models\varphi_1$, so $\rho^{t_0...}\models\varphi$ by Def. \ref{FiniteTraceDef} \textbf{(1i)}.
\par Therefore, $\tau^i =\{t_0,...,T_f\}\setminus \tau^a$ (see \textbf{2i}). \qed  \\

\bottomrule\\
\end{tabular}
\end{table}
\FloatBarrier

\begin{table}[h]
\centering
\begin{tabular}{lp{13.0 cm}}
\multicolumn{2}{l}{\textbf{Not Violated status timeset $\tau^v$}}\\
\toprule
\textbf{If} $\tau^v_1=\emptyset$:&\textbf{Set $\tau^v=\{t_0,...,T_f\}$}\\
\midrule
& $\rho^{t_0...}\models\varphi_1$, so $\rho^{t_0...}\not\models\varphi$ by Def. \ref{FiniteTraceDef} \textbf{(1v)}. Thus $\tau^v=\{t_0,...,T_f\}$. \qed  \\
\midrule
\textbf{Else}:&\textbf{Set} $\tau^v= \emptyset$\\
\midrule
& $\rho^{t_0...}\not\models\varphi_1$, so $\rho^{t_0...}\models\varphi$ by Def. \ref{FiniteTraceDef} and \textbf{no} $t$ meets (1v). Thus $\tau^v=\emptyset$. \qed  \\

\bottomrule\\
\end{tabular}
\end{table}
\FloatBarrier
\end{proof}

\subsection{And}

\begin{table}[h]
\centering
\begin{tabular}{lp{13.0 cm}}
\toprule
\textbf{And} $(and)$&LTL: $\varphi=\varphi_1\wedge\varphi_2$\\
\midrule
Initialize: & Load all $\tau_1^q,\tau_2^q$ of child nodes $\varphi_1, \varphi_2$ on $\rho^{t_0...}$.\\
If $\tau^v_1$ or $\tau^v_2\neq\emptyset$:&Set $\tau^v=\{t_0,...,T_f\}$; $\tau^a,\tau^s,\tau^i=\emptyset$. \\
Else:&Set $\tau^a,\tau^s=\{t_0\}, \tau^i=\{t_0,...,T_f\}\setminus \tau^a$. Set $\tau^v=\emptyset$.\\

\bottomrule\\
\end{tabular}
\caption{\label{tab:andmodule} And module definition.}
\end{table}
\FloatBarrier

\begin{tcolorbox}[colback=gray!10!white,colframe=white!80!black,title=\color{black}{Definition \ref{FiniteTraceDef}: Formula Evaluation on Finite Traces (and)}]
  LTL formula $\varphi$ is true on finite trace $\rho^{t_0...} = (L_{t_0},...,L_{T_f})$, denoted $\rho^{t_0...}\models\varphi$, if $T_0\leq t_0 \leq T_f$ \textbf{and}:
        \par$\rho^{t_0...} \models \varphi_1 \wedge \varphi_2\quad$ \textbf{iff} $\quad\rho^{t_0...}\models\varphi_1$ \textbf{and} $\rho^{t_0...}\models\varphi_2$
\end{tcolorbox}

\begin{tcolorbox}[colback=gray!50!white,colframe=white!60!black,title=Properties of $\varphi$]
\par \textbf{Precondition}: $true$ (Def \ref{preconditionDef})
\par \textbf{Arbitrary suffixes}: all $\rho^{t...}$ such that $t_0<t\leq T_f$ (Def \ref{arbitraryDef})
\begin{itemize}
\item Consider $\varphi = \varphi_1 \wedge \varphi_2$. By Def. \ref{FiniteTraceDef}, $\rho^{t_0...}\models\varphi$ iff ($\rho^{t_0...}\models\varphi_1$ and $\rho^{t_0...}\models\varphi_2$). Thus, $\rho^{t_0...}\models\varphi$ does not depend on whether $\rho^{t...}\models\varphi_j$ where $j=1,2$ for any $t_0<t\leq T_f$, making all $\rho^{t...}$ with $t>t_0$ arbitrary.
\end{itemize}
\end{tcolorbox}

\begin{proof}[Proof: Soundness of And Module] We examine the outputs of the And module status by status. The status definition conditions are labeled \textbf{(1a)}, \textbf{(2a)}, etc. as in the reproduction of Definition \ref{activedef} above.

\begin{table}[h]
\centering
\begin{tabular}{lp{13.0 cm}}
\multicolumn{2}{l}{\textbf{And Active status timeset $\tau^a$}}\\
\toprule
\textbf{If} $\tau^v_1$ \textbf{or} $\tau^v_2\neq\emptyset$:&\textbf{Set} $\tau^a=\emptyset$\\
\midrule
& \par $\bullet$ $\tau^v_j\neq\emptyset$ on $\rho^{t_0...}$ iff $\rho^{t_0...}\not\models\varphi_j$ by Def. \ref{activedef}. Thus ($\tau^v_1$ \textbf{or} $\tau^v_2\neq\emptyset$) iff $\rho^{t_0...}\not\models\varphi_j$ for $j=1$ or $j=2$.
\par $\bullet$ $\rho^{t_0...}\not\models\varphi_j$ for $j=1$ or $j=2$ means that $\rho^{t_0...}\not\models\varphi_1\wedge\varphi_2$ by Def. \ref{FiniteTraceDef}.
\par Thus, (1a) is not met and $\tau^a=\emptyset$.
\qed  \\
\midrule
\textbf{Else}:&\textbf{Set} $\tau^a=\{t_0\}$\\
\midrule
&\par $\bullet$ $\rho^{t_0...}\models\varphi_1$ and $\rho^{t_0...}\models\varphi_2$, which means $\rho^{t_0...}\models\varphi_1\wedge\varphi_2$ by Def. \ref{FiniteTraceDef} \textbf{(1a)}.
\par $\bullet$ $\rho^{t_0...}\models true$ \textbf{(2a)}
\par $\bullet$ All $\rho^{t...}$ are arbitrary except at $t=t_0$ for $t_0\leq t \leq T_f$ \textbf{(3a for $t_0$)}
\par Therefore, \textbf{only $t=t_0$} meets all conditions for active status, and $\tau^a=\{t_0\}$. \qed  \\

\bottomrule\\
\end{tabular}
\end{table}
\FloatBarrier

\begin{table}[h]
\centering
\begin{tabular}{lp{13.0 cm}}
\multicolumn{2}{l}{\textbf{And Satisfied status timeset $\tau^s$}}\\
\toprule
\textbf{If} $\tau^v_1$ \textbf{or} $\tau^v_2\neq\emptyset$:&\textbf{Set} $\tau^s=\emptyset$\\
\midrule
& \par Since $\varphi$ is not active at any $t$, (1s) is not met for any $t$, so $\tau^s=\emptyset$. \qed  \\
\midrule
\textbf{Else}:&\textbf{Set} $\tau^s=\{t_0\}$\\
\midrule
& $\bullet$ $\varphi$ is active at $t_0$ \textbf{(1s for $t_0$)} 
\par $\bullet$ $\varphi$ not active at $t>t_0$ \textbf{(2s for $t_0$)}. For all $t>t_0$, (1s) is not met. 
\par Therefore, \textbf{only} $t=t_0$ meets all conditions for satisfied status, and $\tau^s=\{t_0\}$. \qed  \\

\bottomrule\\
\end{tabular}
\end{table}
\FloatBarrier

\begin{table}[h]
\centering
\begin{tabular}{lp{13.0 cm}}
\multicolumn{2}{l}{\textbf{And Inactive status timeset $\tau^i$}}\\
\toprule
\textbf{If} $\tau^v_1$ \textbf{or} $\tau^v_2\neq\emptyset$:&\textbf{Set} $\tau^i=\emptyset$\\
\midrule
& $\bullet$ Again, $\tau^v_j\neq\emptyset$ on $\rho^{t_0...}$ iff $\rho^{t_0...}\not\models\varphi_j$ by Def. \ref{activedef}, so $\rho^{t_0...}\not\models\varphi_j$ for $j=1$ or $j=2$.
\par $\bullet$ $\rho^{t_0...}\not\models\varphi_j$ for $j=1$ or $j=2$ means that $\rho^{t_0...}\not\models\varphi_1\wedge\varphi_2$ by Def. \ref{FiniteTraceDef}.
\par Thus, (1i) is not met and $\tau^i=\emptyset$. \qed  \\
\midrule
\textbf{Else}:&\textbf{Set} $\tau^i=\{t_0,...,T_f\}\setminus \tau^a$\\
\midrule
& $\bullet$ $\rho^{t_0...}\models\varphi$ by Def. \ref{FiniteTraceDef} \textbf{(1i)}.
\par Therefore, $\tau^i =\{t_0,...,T_f\}\setminus \tau^a$ (see \textbf{2i}). \qed  \\

\bottomrule\\
\end{tabular}
\end{table}
\FloatBarrier

\begin{table}[h]
\centering
\begin{tabular}{lp{13.0 cm}}
\multicolumn{2}{l}{\textbf{And Violated status timeset $\tau^v$}}\\
\toprule
\textbf{If} $\tau^v_1$ \textbf{or} $\tau^v_2\neq\emptyset$:&\textbf{Set} $\tau^v=\{t_0,...,T_f\}$\\
\midrule
& Again, $\rho^{t_0...}\not\models\varphi$ by Def. \ref{FiniteTraceDef} \textbf{(1v)}. Thus $\tau^v=\{t_0,...,T_f\}$. \qed  \\
\midrule
\textbf{Else}:&\textbf{Set} $\tau^v=\emptyset$\\
\midrule
& $\rho^{t_0...}\models\varphi$ by Def. \ref{FiniteTraceDef} and \textbf{no} $t$ meets (1v). Thus $\tau^v=\emptyset$. \qed \\

\bottomrule\\
\end{tabular}
\end{table}
\FloatBarrier
\end{proof}


\subsection{Or}

\begin{table}[h]
\centering
\begin{tabular}{lp{13.0 cm}}
\toprule
\textbf{Or} $(or)$&LTL: $\varphi=\varphi_1\vee\varphi_2$\\
\midrule
Initialize: & Load all $\tau_1^q,\tau_2^q$ of child nodes $\varphi_1, \varphi_2$ on $\rho^{t_0...}$.\\
If $\tau^v_1,\tau^v_2\neq\emptyset$:&Set $\tau^v=\{t_0,...,T_f\}$; $\tau^a,\tau^s,\tau^i=\emptyset$. \\
Else:&Set $\tau^a,\tau^s=\{t_0\}, \tau^i=\{t_0,...,T_f\}\setminus \tau^a$. Set $\tau^v=\emptyset$.\\

\bottomrule\\
\end{tabular}
\caption{\label{tab:ormodule} Or module definition.}
\end{table}
\FloatBarrier

\begin{tcolorbox}[colback=gray!10!white,colframe=white!80!black,title=\color{black}{Definition \ref{FiniteTraceDef}: Formula Evaluation on Finite Traces (or)}]
LTL formula $\varphi$ is true on finite trace $\rho^{t_0...} = (L_{t_0},...,L_{T_f})$, denoted $\rho^{t_0...}\models\varphi$, if $T_0\leq t_0 \leq T_f$ \textbf{and}:
 \par$\rho^{t_0...} \models \varphi_1\vee\varphi_2 \quad$ \textbf{where} $\quad \rho^{t_0...} \models \neg(\neg\varphi_1\wedge\neg\varphi_2).$
 \begin{itemize}
     \item The requirement $\rho^{t_0...}\models\neg\varphi_1\wedge\neg\varphi_2$ is equivalent to requiring $\rho^{t_0...}\not\models\varphi_1$ and $\rho^{t_0...}\not\models\varphi_2$ (by def. of $\wedge$ and $\neg$). This condition, $(\rho^{t_0...}\not\models\varphi_1$ and $\rho^{t_0...}\not\models\varphi_2)$ is false when $(\rho^{t_0...}\models\varphi_1)$ or $(\rho^{t_0...}\models\varphi_2)$ is true. The final condition is thus
 \end{itemize}
 \par $\rho^{t_0...} \models \varphi_1\vee\varphi_2 \quad$ \textbf{iff} $\quad \rho^{t_0...}\models\varphi_1$ \textbf{or} $\rho^{t_0...}\models\varphi_2$
\end{tcolorbox}

\begin{tcolorbox}[colback=gray!50!white,colframe=white!60!black,title=Properties of $\varphi$]
\par \textbf{Precondition}: $true$ (Def \ref{preconditionDef})
\par \textbf{Arbitrary suffixes}: all $\rho^{t...}$ such that $t_0<t\leq T_f$ (Def \ref{arbitraryDef})
\begin{itemize}
\item Consider $\varphi = \varphi_1 \vee \varphi_2$. By Def. \ref{FiniteTraceDef}, $\rho^{t_0...}\models\varphi$ iff ($\rho^{t_0...}\models\varphi_1$ or $\rho^{t_0...}\models\varphi_2$). Thus, $\rho^{t_0...}\models\varphi$ does not depend on whether $\rho^{t...}\models\varphi_j$ where $j=1,2$ for any $t_0<t\leq T_f$, making all $\rho^{t...}$ with $t>t_0$ arbitrary.
\end{itemize}
\end{tcolorbox}

\begin{proof}[Proof: Soundness of Or Module] We examine the outputs of the Or module status by status. The status definition conditions are labeled \textbf{(1a)}, \textbf{(2a)}, etc. as in the reproduction of Definition \ref{activedef} above.

\begin{table}[h]
\centering
\begin{tabular}{lp{13.0 cm}}
\multicolumn{2}{l}{\textbf{Or Active status timeset $\tau^a$}}\\
\toprule
\textbf{If} $\tau^v_1,\tau^v_2\neq\emptyset$:&\textbf{Set} $\tau^a=\emptyset$\\
\midrule
& \par $\bullet$ $\tau^v_j\neq\emptyset$ on $\rho^{t_0...}$ iff $\rho^{t_0...}\not\models\varphi_j$ by Def. \ref{activedef}. Thus $\tau^v_1,\tau^v_2\neq\emptyset$ iff $\rho^{t_0...}\not\models\varphi_j$ for $j=1,2$.
\par $\bullet$ $\rho^{t_0...}\not\models\varphi_j$ for $j=1,2$ means that $\rho^{t_0...}\not\models\varphi_1\vee\varphi_2$ by Def. \ref{FiniteTraceDef}.
\par Thus, (1a) is not met and $\tau^a=\emptyset$.
\qed  \\
\midrule
\textbf{Else}:&\textbf{Set} $\tau^a=\{t_0\}$\\
\midrule
& \par $\bullet$ $\rho^{t_0...}\models\varphi_1$ or $\rho^{t_0...}\models\varphi_2$, which means $\rho^{t_0...}\models\varphi_1\vee\varphi_2$ by Def. \ref{FiniteTraceDef} \textbf{(1a)}.
\par $\bullet$ $\rho^{t_0...}\models true$ \textbf{(2a)}
\par $\bullet$ All $\rho^{t...}$ are arbitrary except at $t=t_0$ for $t_0\leq t \leq T_f$ \textbf{(3a for $t_0$)}
\par Therefore, \textbf{only $t=t_0$} meets all conditions for active status, and $\tau^a=\{t_0\}$. \qed  \\

\bottomrule\\
\end{tabular}
\end{table}
\FloatBarrier

\begin{table}[h]
\centering
\begin{tabular}{lp{13.0 cm}}
\multicolumn{2}{l}{\textbf{Or Satisfied status timeset $\tau^s$}}\\
\toprule
\textbf{If} $\tau^v_1,\tau^v_2\neq\emptyset$:&\textbf{Set} $\tau^s=\emptyset$\\
\midrule
& \par Since $\varphi$ is not active at any $t$, (1s) is not met for any $t$, so $\tau^s=\emptyset$. \qed  \\
\midrule
\textbf{Else}:&\textbf{Set} $\tau^s=\{t_0\}$.\\
\midrule
& \par $\bullet$ $\varphi$ not active at $t>t_0$ \textbf{(2s for $t_0$)}. For all $t>t_0$, (1s) is not met. 
\par Therefore, \textbf{only} $t=t_0$ meets all conditions for satisfied status, and $\tau^s=\{t_0\}$. \qed  \\

\bottomrule\\
\end{tabular}
\end{table}
\FloatBarrier

\begin{table}[h]
\centering
\begin{tabular}{lp{13.0 cm}}
\multicolumn{2}{l}{\textbf{Or Inactive status timeset $\tau^i$}}\\
\toprule
\textbf{If} $\tau^v_1,\tau^v_2\neq\emptyset$:&\textbf{Set} $\tau^i=\emptyset$\\
\midrule
& Again, $\tau^v_j\neq\emptyset$ on $\rho^{t_0...}$ iff $\rho^{t_0...}\not\models\varphi_j$ by Def. \ref{activedef}, so $\rho^{t_0...}\not\models\varphi_j$ for $j=1,2$.
\par $\bullet$ $\rho^{t_0...}\not\models\varphi_j$ for $j=1,2$ means that $\rho^{t_0...}\not\models\varphi_1\vee\varphi_2$ by Def. \ref{FiniteTraceDef}.
\par Thus, (1i) is not met and $\tau^i=\emptyset$. \qed \\
\midrule
\textbf{Else}:&\textbf{Set} $\tau^i=\{t_0,...,T_f\}\setminus \tau^a$\\
\midrule
& $\bullet$ $\rho^{t_0...}\models\varphi$ by Def. \ref{FiniteTraceDef} \textbf{(1i)}.
\par Therefore, $\tau^i =\{t_0,...,T_f\}\setminus \tau^a$ (see \textbf{2i}). \qed  \\

\bottomrule\\
\end{tabular}
\end{table}
\FloatBarrier

\begin{table}[h]
\centering
\begin{tabular}{lp{13.0 cm}}
\multicolumn{2}{l}{\textbf{Or Violated status timeset $\tau^v$}}\\
\toprule
\textbf{If} $\tau^v_1,\tau^v_2\neq\emptyset$:&\textbf{Set} $\tau^v=\{t_0,...,T_f\}$\\
\midrule
&Again, $\rho^{t_0...}\not\models\varphi$ by Def. \ref{FiniteTraceDef} \textbf{(1v)}. Thus $\tau^v=\{t_0,...,T_f\}$. \qed \\
\midrule
\textbf{Else}:&\textbf{Set} $\tau^v=\emptyset$.\\
\midrule
&$\rho^{t_0...}\models\varphi$ by Def. \ref{FiniteTraceDef} and \textbf{no} $t$ meets (1v). Thus $\tau^v=\emptyset$. \qed  \\

\bottomrule\\
\end{tabular}
\end{table}
\FloatBarrier
\end{proof}

\subsection{Implication}

\begin{table}[h]
\centering
\begin{tabular}{lp{13.0 cm}}
\toprule
\textbf{Implication}  $(\rightarrow)$&LTL: $\varphi=\varphi_1\rightarrow\varphi_2$\\
\midrule
Initialize: & Load all $\tau_1^q$ of child node $\varphi_1$ on $\rho^{t_0...}$.\\
If $\tau^v_1\neq\emptyset$:&Set $\tau^a,\tau^s,\tau^v=\emptyset$; $\tau^i=\{t_0,...,T\}\setminus\tau^a.$ \textbf{Exit the current module.}\\
Else:&Load all $\tau_2^q$ of child node $\varphi_2$ on $\rho^{t_0...}$.\\
If $\tau^v_2\neq\emptyset$:&Set $\tau^v=\{t_0,...,T_f\}$; $\tau^a,\tau^s,\tau^i=\emptyset$.\\
Else:&Set $\tau^a, \tau^s=\{t_0\}; \tau^i=\{t_0,...,T_f\}\setminus\tau^a$. Set $\tau^v=\emptyset$.\\
\bottomrule\\
\end{tabular}
\caption{\label{tab:implmodule} Implication module definition.}
\end{table}
\FloatBarrier

\begin{tcolorbox}[colback=gray!10!white,colframe=white!80!black,title=\color{black}{Definition \ref{FiniteTraceDef}: Formula Evaluation on Finite Traces ($\rightarrow$)}]
LTL formula $\varphi$ is true on finite trace $\rho^{t_0...} = (L_{t_0},...,L_{T_f})$, denoted $\rho^{t_0...}\models\varphi$, if $T_0\leq t_0 \leq T_f$ \textbf{and}:
 \par$\rho^{t_0...} \models \varphi_1\rightarrow\varphi_2 \quad$ where $\quad \rho^{t_0...} \models \neg\varphi_1\vee\varphi_2$

\begin{itemize}
    \item  We see that $\rho^{t_0...} \models \neg\varphi_1\vee\varphi_2$ when $\rho^{t_0...}\models\neg\varphi_1$ or $\rho^{t_0...}\models\varphi_2$ (by def. of $\vee$). Thus,
\end{itemize}

  \par$\rho^{t_0...} \models \varphi_1\rightarrow\varphi_2 \quad$ \textbf{iff} $\quad\rho^{t_0...}\not\models\varphi_1$ \textbf{or} $\rho^{t_0...}\models\varphi_2$
\end{tcolorbox}

\begin{tcolorbox}[colback=gray!50!white,colframe=white!60!black,title=Properties of $\varphi$]
\par \textbf{Precondition}: $\varphi_1$ (Def \ref{preconditionDef})
\par \textbf{Arbitrary suffixes}: all $\rho^{t...}$ such that $t_0<t\leq T_f$ (Def \ref{arbitraryDef})
\begin{itemize}
\item Consider $\varphi = \varphi_1 \rightarrow \varphi_2$. By Def. \ref{FiniteTraceDef}, $\rho^{t_0...}\models\varphi$ iff ($\rho^{t_0...}\not\models\varphi_1$ or $\rho^{t_0...}\models\varphi_2$). Thus, $\rho^{t_0...}\models\varphi$ does not depend on whether $\rho^{t...}\models\varphi_j$ where $j=1,2$ for any $t_0<t\leq T_f$, making all $\rho^{t...}$ with $t>t_0$ arbitrary.
\end{itemize}
\end{tcolorbox}

\begin{proof}[Proof: Soundness of Implication Module] We examine the outputs of the Implication module status by status. The status definition conditions are labeled \textbf{(1a)}, \textbf{(2a)}, etc. as in the reproduction of Definition \ref{activedef} above.

\begin{table}[h]
\centering
\begin{tabular}{lp{13.0 cm}}
\multicolumn{2}{l}{\textbf{Implication Active status timeset $\tau^a$}}\\
\toprule
\textbf{If} $\tau^v_1\neq\emptyset$:&\textbf{Set} $\tau^a=\emptyset$ \& \textbf{exit the current module}\\
\midrule
& \par $\tau^v_1\neq\emptyset$ on $\rho^{t_0...}$ iff $\rho^{t_0...}\not\models\varphi_1$ by Def. \ref{activedef}. Thus (2a) is \textbf{not} met, so $\tau^a=\emptyset$. \qed \\
\midrule
\textbf{Else}:&\textbf{Load} all $\tau_2^q$ of child node $\varphi_2$ on $\rho^{t_0...}$\\
\midrule
\textbf{If} $\tau^v_2\neq\emptyset$:&\textbf{Set} $\tau^a=\emptyset$\\
\midrule
&$\bullet$ $\tau^v_2\neq\emptyset$ on $\rho^{t_0...}$ iff $\rho^{t_0...}\not\models\varphi_2$ by Def. \ref{activedef}; we have $\rho^{t_0...}\models\varphi_1$ \textbf{and} $\rho^{t_0...}\not\models\varphi_2$.
\par Thus, $\rho^{t_0...}\not\models\varphi_1\rightarrow\varphi_2$ by Def. \ref{FiniteTraceDef} and (1a) is \textbf{not} met, so $\tau^a=\emptyset$. \qed \\
\midrule
\textbf{Else}:&\textbf{Set} $\tau^a=\{t_0\}$\\
\midrule
& $\bullet$ We have $\rho^{t_0...}\models\varphi_1$ \textbf{and} $\rho^{t_0...}\models\varphi_2$, so $\rho^{t_0...}\models\varphi$ by Def. \ref{FiniteTraceDef} \textbf{(1a)}.
\par $\bullet$ $\rho^{t_0...}\models\varphi_1$ meets precondition \textbf{(2a)}.
\par $\bullet$ All $\rho^{t...}$ are arbitrary except at $t=t_0$ for $t_0\leq t \leq T_f$ \textbf{(3a for $t_0$)}.
\par Therefore, \textbf{only $t=t_0$} meets all conditions for active status, and $\tau^a=\{t_0\}$. \qed \\
\bottomrule
\end{tabular}
\end{table}
\FloatBarrier

\begin{table}[h]
\centering
\begin{tabular}{lp{13.0 cm}}
\multicolumn{2}{l}{\textbf{Implication Satisfied status timeset $\tau^s$}}\\
\toprule
\textbf{If} $\tau^v_1\neq\emptyset$:&\textbf{Set} $\tau^s=\emptyset$ \& \textbf{exit the current module}\\
\midrule
& \par Since $\varphi$ is not active at any $t$, (1s) is not met for any $t$, so $\tau^s=\emptyset$. \qed \\
\midrule
\textbf{Else}:&\textbf{Load} all $\tau_2^q$ of child node $\varphi_2$ on $\rho^{t_0...}$.\\
\midrule
\textbf{If} $\tau^v_2\neq\emptyset$:&\textbf{Set} $\tau^s=\emptyset$.\\
\midrule
& \par Since $\varphi$ is not active at any $t$, (1s) is not met for any $t$, so $\tau^s=\emptyset$. \qed \\
\midrule
\textbf{Else}:&\textbf{Set} $\tau^s=\{t_0\}$\\
\midrule
& \par $\bullet$ $\varphi$ not active at $t>t_0$ \textbf{(2s for $t_0$)}. For all $t>t_0$, (1s) is not met. 
\par Therefore, \textbf{only} $t=t_0$ meets all conditions for satisfied status, and $\tau^s=\{t_0\}$. \qed \\
\bottomrule
\end{tabular}
\end{table}
\FloatBarrier

\begin{table}[h]
\centering
\begin{tabular}{lp{13.0 cm}}
\multicolumn{2}{l}{\textbf{Implication Inactive status timeset $\tau^i$}}\\
\toprule
\textbf{If} $\tau^v_1\neq\emptyset$:&\textbf{Set} $\tau^i=\{t_0,...,T\}\setminus\tau^a$ \& \textbf{exit the current module}\\
\midrule
& $\bullet$ $\tau^v_1\neq\emptyset$ iff $\rho^{t_0...}\not\models\varphi$ by Def. \ref{activedef}. Then $\rho^{t_0...}\models\neg\varphi_1\vee\varphi_2$ by Def. \ref{FiniteTraceDef} \textbf{(1i)}.
\par Therefore, $\tau^i =\{t_0,...,T_f\}\setminus \tau^a$ \textbf{(2i)}. \qed  \\
\midrule
\textbf{Else}:&\textbf{Load} all $\tau_2^q$ of child node $\varphi_2$ on $\rho^{t_0...}$.\\
\midrule
\textbf{If} $\tau^v_2\neq\emptyset$:&\textbf{Set} $\tau^i=\emptyset$.\\
\midrule
&$\bullet$ Again, $\tau^v_2\neq\emptyset$ on $\rho^{t_0...}$ iff $\rho^{t_0...}\not\models\varphi_2$ by Def. \ref{activedef}.
\par Thus $\rho^{t_0...}\not\models\varphi_1\rightarrow\varphi_2$ by Def. \ref{FiniteTraceDef} and (1i) is \textbf{not} met, so $\tau^i=\emptyset$. \qed \\
\midrule
\textbf{Else}:&\textbf{Set} $\tau^i=\{t_0,...,T_f\}\setminus\tau^a$\\
\midrule
& $\bullet$ We have $\rho^{t_0...}\models\varphi_1$ \textbf{and} $\rho^{t_0...}\models\varphi_2$, so $\rho^{t_0...}\models\varphi$ by Def. \ref{FiniteTraceDef} \textbf{(1i)}.
\par Therefore, $\tau^i =\{t_0,...,T_f\}\setminus \tau^a$ (see \textbf{2i}). \qed  \\
\bottomrule
\end{tabular}
\end{table}
\FloatBarrier

\begin{table}[h]
\centering
\begin{tabular}{lp{13.0 cm}}
\multicolumn{2}{l}{\textbf{Implication Violated status timeset $\tau^v$}}\\
\toprule
\textbf{If} $\tau^v_1\neq\emptyset$:&\textbf{Set} $\tau^v=\emptyset$ \& \textbf{exit the current module}\\
\midrule
&$\rho^{t_0...}\models\varphi$ by Def. \ref{FiniteTraceDef} and \textbf{no} $t$ meets (1v). Thus $\tau^v=\emptyset$. \qed \\
\midrule
\textbf{Else}:&\textbf{Load} all $\tau_2^q$ of child node $\varphi_2$ on $\rho^{t_0...}$.\\
\midrule
\textbf{If} $\tau^v_2\neq\emptyset$:&\textbf{Set} $\tau^v=\{t_0,...,T_f\}$\\
\midrule
&Again, $\rho^{t_0...}\not\models\varphi$ by Def. \ref{FiniteTraceDef} \textbf{(1v)}. Thus $\tau^v=\{t_0,...,T_f\}$. \qed \\
\midrule
\textbf{Else}:&\textbf{Set} $\tau^v=\emptyset$\\
\midrule
&We again have $\rho^{t_0...}\models\varphi$ by Def. \ref{FiniteTraceDef}, so \textbf{no} $t$ meets (1v). Thus $\tau^v=\emptyset$. \qed \\
\bottomrule
\end{tabular}
\end{table}
\FloatBarrier

\end{proof}

\section{Temporal modules}

\subsection{Next}

\begin{table}[h]
\centering
\begin{tabular}{lp{13.0 cm}}
\toprule
\textbf{Next}  $(X)$&LTL: $\varphi=\mathbf{X}\varphi_1$\\
\midrule
Initialize: & Load all $\tau_1^q$ of child node $\varphi_1$ at $t_0^1=t_1$. \\
If $\tau_1^v = \emptyset$:& Set $\tau^a = \{t_0, t_1\}$; $\tau^s = \{t_1\}$, $\tau^i = \{t_0,...,T_f\}\setminus \tau^a$. Set $\tau^v = \emptyset$.\\
Else:& Set $\tau^v = \{t_0,...,T_f\}$; $\tau^a,\tau^s,\tau^i=\emptyset$.\\

\bottomrule\\
\end{tabular}
\caption{\label{tab:Xmodule} Next module definition.}
\end{table}
\FloatBarrier

\begin{tcolorbox}[colback=gray!10!white,colframe=white!80!black,title=\color{black}{Definition \ref{FiniteTraceDef}: Formula Evaluation on Finite Traces (X)}]
LTL formula $\varphi$ is true on finite trace $\rho^{t_0...} = (L_{t_0},...,L_{T_f})$, denoted $\rho^{t_0...}\models\varphi$, if $T_0\leq t_0 \leq T_f$ \textbf{and}:
 \par $\rho^{t_0...} \models \mathbf{X}\varphi\quad$ \textbf{iff} $\quad \rho^{t_0+1...}\models\varphi$ \textbf{and} $t_0<T_f$

\end{tcolorbox}

\begin{tcolorbox}[colback=gray!50!white,colframe=white!60!black,title=Properties of $\varphi$]
\par \textbf{Precondition}: $true$ (Def \ref{preconditionDef})
\par \textbf{Arbitrary suffixes}: all $\rho^{t...}$ such that $t_0+1<t\leq T_f$ (Def \ref{arbitraryDef})
\begin{itemize}
\item Consider $\varphi = \mathbf{X}\varphi_1$. By Def. \ref{FiniteTraceDef}, $\rho^{t_0...}\models\varphi$ iff $\rho^{t_0+1...}\models\varphi_1$. Thus, $\rho^{t_0...}\models\varphi$ does not depend on whether $\rho^{t...}\models\varphi_1$ for any $t_0+1<t\leq T_f$. Taking $t_1:=t_0+1$, all $\rho^{t...}$ with $t>t_1$ are then arbitrary.
\end{itemize}
\end{tcolorbox}

\begin{proof}[Proof: Soundness of Next Module] We examine the outputs of the Next module status by status. The status definition conditions are labeled \textbf{(1a)}, \textbf{(2a)}, etc. as in the reproduction of Definition \ref{activedef} above.

\begin{table}[h]
\centering
\begin{tabular}{lp{13.0 cm}}
\multicolumn{2}{l}{\textbf{Next Active status timeset $\tau^a$}}\\
\toprule
\textbf{If} $\tau_1^v = \emptyset$:& \textbf{Set} $\tau^a = \{t_0, t_1\}$\\
\midrule
& $\bullet$ $\tau^v_1=\emptyset$ on $\rho^{t_1...}$ iff $\rho^{t_1...}\models\varphi_1$ by Def. \ref{activedef}. Then $\rho^{t_0...}\models\mathbf{X}\varphi_1$ by Def. \ref{FiniteTraceDef} \textbf{(1a)}.
\par $\bullet$ $\rho^{t_0...}\models true$ \textbf{(2a)}
\par $\bullet$ All $\rho^{t...}$ are arbitrary except when $t\leq t_1$ for $t_0\leq t \leq T_f$ \textbf{(3a for $t_0,t_1$)}.
\par Therefore, \textbf{only $t=t_0,t_1$} meets all conditions for active status, and $\tau^a=\{t_0\}$. \qed \\
\midrule
\textbf{Else}:& \textbf{Set} $\tau^a=\emptyset$\\
\midrule
& $\bullet$ $\tau^v_1\neq\emptyset$ on $\rho^{t_1...}$ iff $\rho^{t_1...}\not\models\varphi_1$ by Def. \ref{activedef}. 
\par Thus $\rho^{t_0...}\not\models\mathbf{X}\varphi_1$ by Def. \ref{FiniteTraceDef}, (1a) is not met and $\tau^a=\emptyset$. \qed \\
\bottomrule
\end{tabular}
\end{table}
\FloatBarrier

\begin{table}[h]
\centering
\begin{tabular}{lp{13.0 cm}}
\multicolumn{2}{l}{\textbf{Next Satisfied status timeset $\tau^s$}}\\
\toprule
\textbf{If} $\tau_1^v = \emptyset$:& \textbf{Set} $\tau^s = \{t_1\}$\\
\midrule
& $\bullet$ $\varphi$ is active at $t_0,t_1$ \textbf{(1s for $t_0,t_1$)} and not active at $t>t_1$ \textbf{(2s for $t_1$)}. \par $\bullet$ For all $t>t_1$, (1s) is not met; for $t_0$, (2s) is not met.
\par Therefore, \textbf{only} $t=t_1$ meets all conditions, and $\tau^s=\{t_1\}$. \qed \\
\midrule
\textbf{Else}:& \textbf{Set} $\tau^s=\emptyset$.\\
\midrule
&\par Since $\varphi$ is not active at any $t$, (1s) is not met for any $t$, so $\tau^s=\emptyset$. \qed \\
\bottomrule
\end{tabular}
\end{table}
\FloatBarrier

\begin{table}[h]
\centering
\begin{tabular}{lp{13.0 cm}}
\multicolumn{2}{l}{\textbf{Next Inactive status timeset $\tau^i$}}\\
\toprule
\textbf{If} $\tau_1^v = \emptyset$:& \textbf{Set} $\tau^i = \{t_0,...,T_f\}\setminus \tau^a$\\
\midrule
& $\bullet$ $\tau^v_1=\emptyset$ on $\rho^{t_1...}$ iff $\rho^{t_1...}\models\varphi_1$ by Def. \ref{activedef}. Then $\rho^{t_0...}\models\mathbf{X}\varphi_1$ by Def. \ref{FiniteTraceDef} \textbf{(1i)}.
\par Therefore, $\tau^i =\{t_0,...,T_f\}\setminus \tau^a$ (see \textbf{2i}). \qed \\
\midrule
\textbf{Else}:& \textbf{Set} $\tau^i=\emptyset$\\
\midrule
&$\rho^{t_1...}\not\models\varphi_1$, so $\rho^{t_0...}\not\models\mathbf{X}\varphi_1$ by Def. \ref{FiniteTraceDef} and (1i) is \textbf{not} met. Thus $\tau^i=\emptyset$.\qed \\
\bottomrule
\end{tabular}
\end{table}
\FloatBarrier

\begin{table}[h]
\centering
\begin{tabular}{lp{13.0 cm}}
\multicolumn{2}{l}{\textbf{Next Violated status timeset $\tau^v$}}\\
\toprule
\textbf{If} $\tau_1^v = \emptyset$:& \textbf{Set} $\tau^v = \emptyset$\\
\midrule
&$\rho^{t_0...}\models\varphi$ by Def. \ref{FiniteTraceDef} and \textbf{no} $t$ meets (1v). Thus $\tau^v=\emptyset$. \qed \\
\midrule
\textbf{Else}:& \textbf{Set} $\tau^v = \{t_0,...,T_f\}$\\
\midrule
&Again, $\rho^{t_0...}\not\models\varphi$ by Def. \ref{FiniteTraceDef} \textbf{(1v)}. Thus $\tau^v=\{t_0,...,T_f\}$. \qed \\
\bottomrule
\end{tabular}
\end{table}
\FloatBarrier

\end{proof}


\subsection{Eventual}

\begin{table}[h]
\centering
\begin{tabular}{lp{13.0 cm}}
\toprule
\textbf{Eventual} $(F)$&LTL: $\varphi=\mathbf{F}\varphi_1$\\
\midrule
For $t_0^1=t_0...T_f$:& Load all $\tau_1^q$ of child node $\varphi_1$ on $\rho^{t_0^1...}$. Continue until $t_0^1=T_f$ or $\tau^v_1 = \emptyset$.\\
If $\tau_1^v=\emptyset$:&Set $\tau^a = \{t_0,...,t_0^1\}$; $\tau^s = \{t_0^1\}$; $\tau^i=\{t_0,...,T_f\}\setminus\tau^a$. Set $\tau^v=\emptyset$. \textbf{Exit the current module.}\\
 Else:&Set $\tau^v = \{t_0,...,T_f\}$; $\tau^a,\tau^s,\tau^i=\emptyset$.\\

\bottomrule\\
\end{tabular}
\caption{\label{tab:Fmodule} Eventual module definition.}
\end{table}
\FloatBarrier

\begin{tcolorbox}[colback=gray!10!white,colframe=white!80!black,title=\color{black}{Definition \ref{FiniteTraceDef}: Formula Evaluation on Finite Traces (F)}]
LTL formula $\varphi$ is true on finite trace $\rho^{t_0...} = (L_{t_0},...,L_{T_f})$, denoted $\rho^{t_0...}\models\varphi$, if $T_0\leq t_0 \leq T_f$ \textbf{and}:
 \par $\rho^{t_0...} \models \mathbf{F}\varphi \quad$ where $\quad \rho^{t_0...} \models true\ \mathbf{U}\varphi$

\begin{itemize}
    \item By definition, $\rho^{t_0...}\models true\  \mathbf{U}\varphi_2\quad$ iff $\quad \exists i\geq t_0$ s.t. $\rho^{i...}\models\varphi_2$ and $\rho^{k...} \models true$ for all $t_0\leq k < i$ and $i\leq T_f$. Clearly $\rho^{k...} \models true$ trivially, and thus
\end{itemize}

  \par $\rho^{t_0...} \models \mathbf{F}\varphi \quad$ \textbf{iff} $\quad \exists i\geq t_0$ s.t. $\rho^{i...}\models\varphi$ for $i\leq T_f$

\end{tcolorbox}

\begin{tcolorbox}[colback=gray!50!white,colframe=white!60!black,title=Properties of $\varphi$]
\par \textbf{Precondition}: $true$ (Def \ref{preconditionDef})
\par \textbf{Arbitrary suffixes}: all $\rho^{t...}$ such that $t'<t\leq T_f$, where $t'=\min\{i\ |\ \rho^{i...}\models\varphi_1\}$ (Def \ref{arbitraryDef})
\begin{itemize}
\item Consider $\varphi = \mathbf{F}\varphi_1$. By Def. \ref{FiniteTraceDef}, $\rho^{t_0...}\models\varphi$ iff $\rho^{i...}\models\varphi_1$ for some $i\geq t_0$. Thus, $\rho^{t_0...}\models\varphi$ does not depend on whether $\rho^{t...}\models\varphi_1$ for any $i<t\leq T_f$. Taking $t'=\min\{i\ |\ \rho^{i...}\models\varphi_1\}$, we see that $(\rho^{t'...}\models\varphi_1) \Rightarrow (\rho^{t_0...}\models\varphi)$ regardless of $\rho^{t...}\models\varphi_1$ for any $t>t'$. Thus, all $\rho^{t...}$ with $t'<t\leq T_f$ are arbitrary.
\end{itemize}
\end{tcolorbox}

\begin{proof}[Proof: Soundness of Eventual Module] We examine the outputs of the Eventual module status by status. The status definition conditions are labeled \textbf{(1a)}, \textbf{(2a)}, etc. as in the reproduction of Definition \ref{activedef} above.

\begin{table}[h]
\centering
\begin{tabular}{lp{13.0 cm}}
\multicolumn{2}{l}{\textbf{Eventual Active status timeset $\tau^a$}}\\
\toprule
\multicolumn{2}{l}{(For $t_0^1=t_0...T_f$: Load all $\tau_1^q$ of child node $\varphi_1$ on $\rho^{t_0^1...}$. Continue until $t_0^1=T_f$ or $\tau^v_1 = \emptyset$.)}\\
\midrule
\textbf{If} $\tau_1^v=\emptyset$:&\textbf{Set} $\tau^a = \{t_0,...,t_0^1\}$ \& \textbf{exit the current module}\\
\midrule
& $\bullet$ $\tau^v_1=\emptyset$ on $\rho^{t_0^1...}$ iff $\rho^{t_0^1...}\models\varphi_1$ by Def. \ref{activedef}. Then $\rho^{t_0...}\models\mathbf{F}\varphi_1$ by Def. \ref{FiniteTraceDef} \textbf{(1a)}.
\par $\bullet$ $\rho^{t_0...}\models true$ \textbf{(2a)}
\par $\bullet$ All $\rho^{t_0^1...}$ are arbitrary where $t'<t_0^1\leq T_f$, $t' = \min\{i\ |\ \rho^{i...}\models\varphi_1\}$. Since $t_0^1$ is strictly increasing, the first $t_0^1$ s.t. $\rho^{t_0^1...}\models\varphi_1$ is equal to $\min\{i\ |\ \rho^{i...}\models\varphi_1\}$. 
\par Thus, $\rho^{t...}$ is arbitrary only for $t>t_0^1$ at exit \textbf{(3a for $t_0,...,t_0^1$)}, so $\tau^a=\{t_0,...,t_0^1\}$. \qed  \\
\midrule
\textbf{Else}:&\textbf{Set} $\tau^a=\emptyset$.\\
\midrule
& $\bullet$ $\tau^v_1\neq\emptyset$ on $\rho^{t_0^1...}$ iff $\rho^{t_0^1...}\not\models\varphi_1$ by Def. \ref{activedef}. 
\par Thus, if $\tau^v_1\neq\emptyset$ on all $\rho^{t_0^1...}$, $\rho^{t_0...}\not\models\mathbf{F}\varphi_1$ by Def. \ref{FiniteTraceDef}, (1a) is not met and $\tau^a=\emptyset$. \qed \\
\bottomrule
\end{tabular}
\end{table}
\FloatBarrier

\begin{table}[h]
\centering
\begin{tabular}{lp{13.0 cm}}
\multicolumn{2}{l}{\textbf{Eventual Satisfied status timeset $\tau^s$}}\\
\toprule
\multicolumn{2}{l}{(For $t_0^1=t_0...T_f$: Load all $\tau_1^q$ of child node $\varphi_1$ on $\rho^{t_0^1...}$. Continue until $t_0^1=T_f$ or $\tau^v_1 = \emptyset$.)}\\
\midrule
\textbf{If} $\tau_1^v=\emptyset$:&\textbf{Set} $\tau^s = \{t_0^1\}$ \& \textbf{exit the current module}\\
\midrule
& $\bullet$ $\varphi$ is active at $t_0,...,t_0^1$ \textbf{(1s for $t_0,...,t_0^1$)} and not active at $t>t_0^1$ \textbf{(2s for $t_0^1$)}. 
\par $\bullet$ For all $t>t_0^1$, (1s) is not met; for $t<t_0^1$, (2s) is not met.
\par Therefore, \textbf{only} $t_0^1$ meets all conditions at exit, and $\tau^s=\{t_0^1\}$. \qed \\
\midrule
\textbf{Else}:&\textbf{Set} $\tau^s=\emptyset$.\\
\midrule
&\par Since $\varphi$ is not active at any $t$, (1s) is not met for any $t$, so $\tau^s=\emptyset$. \qed \\
\bottomrule
\end{tabular}
\end{table}
\FloatBarrier

\begin{table}[h]
\centering
\begin{tabular}{lp{13.0 cm}}
\multicolumn{2}{l}{\textbf{Eventual Inactive status timeset $\tau^i$}}\\
\toprule
\multicolumn{2}{l}{(For $t_0^1=t_0...T_f$: Load all $\tau_1^q$ of child node $\varphi_1$ on $\rho^{t_0^1...}$. Continue until $t_0^1=T_f$ or $\tau^v_1 = \emptyset$.)}\\
\midrule
\textbf{If} $\tau_1^v=\emptyset$:&\textbf{Set} $\tau^i=\{t_0,...,T_f\}\setminus\tau^a$ \& \textbf{exit the current module}\\
\midrule
& $\bullet$ $\tau^v_1=\emptyset$ on $\rho^{t_0^1...}$ iff $\rho^{t_0^1...}\models\varphi_1$ by Def. \ref{activedef}. Then $\rho^{t_0...}\models\mathbf{F}\varphi_1$ by Def. \ref{FiniteTraceDef} \textbf{(1i)}.
\par Therefore, $\tau^i =\{t_0,...,T_f\}\setminus \tau^a$ (see \textbf{2i}). \qed \\
\midrule
\textbf{Else}:&\textbf{Set} $\tau^i=\emptyset$.\\
\midrule
& $\rho^{t_0^1...}\not\models\varphi_1$ for any $t_0^1$, so $\rho^{t_0...}\not\models\mathbf{F}\varphi_1$ by Def. \ref{FiniteTraceDef} and (1i) is \textbf{not} met. Thus $\tau^i=\emptyset$.\qed \\
\bottomrule
\end{tabular}
\end{table}
\FloatBarrier

\begin{table}[h]
\centering
\begin{tabular}{lp{13.0 cm}}
\multicolumn{2}{l}{\textbf{Eventual Violated status timeset $\tau^v$}}\\
\toprule
\multicolumn{2}{l}{(For $t_0^1=t_0...T_f$: Load all $\tau_1^q$ of child node $\varphi_1$ on $\rho^{t_0^1...}$. Continue until $t_0^1=T_f$ or $\tau^v_1 = \emptyset$.)}\\
\midrule
\textbf{If} $\tau_1^v=\emptyset$:&\textbf{Set} $\tau^v=\emptyset$ \& \textbf{exit the current module}\\
\midrule
&$\rho^{t_0...}\models\varphi$ by Def. \ref{FiniteTraceDef} and \textbf{no} $t$ meets (1v). Thus $\tau^v=\emptyset$. \qed \\
\midrule
\textbf{Else}:&\textbf{Set} $\tau^v = \{t_0,...,T_f\}$\\
\midrule
&Again, $\rho^{t_0...}\not\models\varphi$ by Def. \ref{FiniteTraceDef} \textbf{(1v)}. Thus $\tau^v=\{t_0,...,T_f\}$. \qed \\
\bottomrule
\end{tabular}
\end{table}
\FloatBarrier

\end{proof}

\subsection{Global}

\begin{table}[h]
\centering
\begin{tabular}{lp{13.0 cm}}
\toprule
\textbf{Global} $(G)$&LTL: $\varphi=\mathbf{G}\varphi_1$\\
\midrule
For $t_0^1=t_0...T_f$:& Load all $\tau^q_1$ of child node $\varphi_1$ on $\rho^{t_0^1...}$. \\
&Continue until $t_0^1=T_f$ or $\tau_1^v \neq \emptyset$.\\
If $\tau_1^v\neq\emptyset$:&Set $\tau^v = \{t_0,...,T_f\}$; $\tau^a,\tau^s,\tau^i=\emptyset$. \textbf{Exit the current module.}\\
 Else:&Set $\tau^a=\{t_0,...,T_f\}; \tau^s=\{T_f\};\tau^i,\tau^v=\emptyset$.\\

\bottomrule\\
\end{tabular}
\caption{\label{tab:Gmodule} Global module definition.}
\end{table}
\FloatBarrier

\begin{tcolorbox}[colback=gray!10!white,colframe=white!80!black,title=\color{black}{Definition \ref{FiniteTraceDef}: Formula Evaluation on Finite Traces (G)}]
LTL formula $\varphi$ is true on finite trace $\rho^{t_0...} = (L_{t_0},...,L_{T_f})$, denoted $\rho^{t_0...}\models\varphi$, if $T_0\leq t_0 \leq T_f$ \textbf{and}:
 \par$\rho^{t_0...} \models \mathbf{G}\varphi \quad$ where $\quad \rho^{t_0...} \models \neg\mathbf{F}\neg\varphi$

\begin{itemize}
    \item We have that $\rho^{t_0...}\models\neg\mathbf{F}\psi$ iff $\not\exists i\geq t_0$ s.t. $\rho^{i...}\models\varphi$ for $i\leq T_f$. If this is true, it must be true that $\rho^{i...}\not\models\psi$ for all $i$ where $t_0\leq i\leq T_f$. Equivalently, $\rho^{i...}\models\neg\psi$ for all $t_0\leq i\leq T_f$. 
    \item Taking $\psi = \neg\varphi,$
\end{itemize}
 
 \par $\rho^{t_0...} \models \mathbf{G}\varphi \quad$ \textbf{iff} $\quad \rho^{i...}\models\varphi$ for all $i$ where $t_0\leq i\leq T_f$

\end{tcolorbox}

\begin{tcolorbox}[colback=gray!50!white,colframe=white!60!black,title=Properties of $\varphi$]
\par \textbf{Precondition}: $true$ (Def \ref{preconditionDef})
\par \textbf{Arbitrary suffixes}: none (Def \ref{arbitraryDef})
\begin{itemize}
\item Consider $\varphi = \mathbf{G}\varphi_1$. By Def. \ref{FiniteTraceDef}, $\rho^{t_0...}\models\varphi$ iff $\rho^{i...}\models\varphi_1$ for all $t_0\leq i\leq T_f$. Thus, no $\rho^{t...}$ with $t_0\leq t\leq T_f$ is arbitrary.
\end{itemize}
\end{tcolorbox}

\begin{proof}[Proof: Soundness of Global Module] We examine the outputs of the Global module status by status. The status definition conditions are labeled \textbf{(1a)}, \textbf{(2a)}, etc. as in the reproduction of Definition \ref{activedef} above.

\begin{table}[h]
\centering
\begin{tabular}{lp{13.0 cm}}
\multicolumn{2}{l}{\textbf{Global Active status timeset $\tau^a$}}\\
\toprule
\multicolumn{2}{l}{(For $t_0^1=t_0...T_f$: Load all $\tau^q_1$ of child node $\varphi_1$ on $\rho^{t_0^1...}$. Continue until $t_0^1=T_f$ or $\tau_1^v \neq \emptyset$.)}\\
\midrule
\textbf{If} $\tau_1^v\neq\emptyset$:&\textbf{Set} $\tau^a=\emptyset$ \& \textbf{exit the current module}\\
\midrule
& $\bullet$ $\tau^v_1\neq\emptyset$ on $\rho^{t_0^1...}$ iff $\rho^{t_0^1...}\not\models\varphi_1$ by Def. \ref{activedef}.
\par Thus, if $\tau^v_1\neq\emptyset$ on any $\rho^{t_0^1...}$, $\rho^{t_0...}\not\models\mathbf{G}\varphi_1$ by Def. \ref{FiniteTraceDef}, (1a) is not met and $\tau^a=\emptyset$. \qed \\
\midrule
\textbf{Else}:&\textbf{Set} $\tau^a=\{t_0,...,T_f\}$\\
\midrule
&$\bullet$ $\tau^v_1=\emptyset$ on $\rho^{t_0^1...}$ iff $\rho^{t_0^1...}\models\varphi_1$ by Def. \ref{activedef}. Since $\tau^v_1=\emptyset$ on all $\rho^{t^1_0...}$, $\rho^{t_0...}\models\mathbf{G}\varphi_1$ by Def. \ref{FiniteTraceDef} \textbf{(1a)}.
\par $\bullet$ $\rho^{t_0...}\models true$ \textbf{(2a)}
\par $\bullet$ $\rho^{t...}$ is not arbitrary for any $t$ \textbf{(3a for $t_0\leq t \leq T_f$)}.
\par Thus, all $t$ meet all criteria and $\tau^a = \{t_0,...,T_f\}$. \qed 
\\
\bottomrule
\end{tabular}
\end{table}
\FloatBarrier

\begin{table}[h]
\centering
\begin{tabular}{lp{13.0 cm}}
\multicolumn{2}{l}{\textbf{Global Satisfied status timeset $\tau^s$}}\\
\toprule
\multicolumn{2}{l}{(For $t_0^1=t_0...T_f$: Load all $\tau^q_1$ of child node $\varphi_1$ on $\rho^{t_0^1...}$. Continue until $t_0^1=T_f$ or $\tau_1^v \neq \emptyset$.)}\\
\midrule
\textbf{If} $\tau_1^v\neq\emptyset$:&\textbf{Set} $\tau^s=\emptyset$ \& \textbf{exit the current module}\\
\midrule
& Since $\varphi$ is not active at any $t$, (1s) is not met for any $t$, so $\tau^s=\emptyset$. \qed \\
\midrule
\textbf{Else}:&\textbf{Set} $\tau^s=\{T_f\}$\\
\midrule
& $\bullet$ $\varphi$ is active on all $t$ \textbf{(1s for $t_0\leq t \leq T_f$)}, \textbf{(2s for $T_f$)}.
\par Therefore, \textbf{only} $T_f$ meets all conditions at exit, and $\tau^s=\{T_f\}$. \qed \\
\bottomrule
\end{tabular}
\end{table}
\FloatBarrier

\begin{table}[h]
\centering
\begin{tabular}{lp{13.0 cm}}
\multicolumn{2}{l}{\textbf{Global Inactive status timeset $\tau^i$}}\\
\toprule
\multicolumn{2}{l}{(For $t_0^1=t_0...T_f$: Load all $\tau^q_1$ of child node $\varphi_1$ on $\rho^{t_0^1...}$. Continue until $t_0^1=T_f$ or $\tau_1^v \neq \emptyset$.)}\\
\midrule
\textbf{If} $\tau_1^v\neq\emptyset$:&\textbf{Set} $\tau^i=\emptyset$ \& \textbf{exit the current module}\\
\midrule
& $\rho^{t_0^1...}\not\models\varphi_1$ for some $t_0^1$, so $\rho^{t_0...}\not\models\mathbf{G}\varphi_1$ by Def. \ref{FiniteTraceDef} (1i \textbf{not} met). Thus $\tau^i=\emptyset$.\qed \\
\midrule
\textbf{Else}:&\textbf{Set} $\tau^i=\emptyset$.\\
\midrule
& $\varphi$ is active for all $t$ (2i not met for $t_0\leq t \leq T_f$). Thus $\tau^i=\emptyset$.\qed \\
\bottomrule
\end{tabular}
\end{table}
\FloatBarrier

\begin{table}[h]
\centering
\begin{tabular}{lp{13.0 cm}}
\multicolumn{2}{l}{\textbf{Global Violated status timeset $\tau^v$}}\\
\toprule
\multicolumn{2}{l}{(For $t_0^1=t_0...T_f$: Load all $\tau^q_1$ of child node $\varphi_1$ on $\rho^{t_0^1...}$. Continue until $t_0^1=T_f$ or $\tau_1^v \neq \emptyset$.)}\\
\midrule
\textbf{If} $\tau_1^v\neq\emptyset$:&\textbf{Set} $\tau^v = \{t_0,...,T_f\}$ \& \textbf{exit the current module}\\
\midrule
&$\rho^{t_0...}\not\models\varphi$ by Def. \ref{FiniteTraceDef} \textbf{(1v)}. Thus $\tau^v=\{t_0,...,T_f\}$. \qed \\
\midrule
\textbf{Else}:&\textbf{Set} $\tau^v=\emptyset$\\
\midrule
&Again, $\rho^{t_0...}\models\varphi$ by Def. \ref{FiniteTraceDef} and \textbf{no} $t$ meets (1v). Thus $\tau^v=\emptyset$. \qed \\
\bottomrule
\end{tabular}
\end{table}
\FloatBarrier

\end{proof}


\subsection{Until}

\begin{table}[h]
\centering
\begin{tabular}{lp{13.0 cm}}
\toprule 
 \textbf{Until} $(U)$&LTL: $\varphi=\varphi_1\mathbf{U}\varphi_2$\\
 \midrule
For $t'=t_0...T_f$:& Load all $\tau_1^q, \tau_2^q$ of child nodes $\varphi_1$ and $\varphi_2$ on $\rho^{t'...}$. Continue until $\tau_2^v = \emptyset$ or $\tau^v_1\neq\emptyset$ or $t'=T_f$.\\
If $\tau_2^v = \emptyset$:&Set $\tau^a = \{t_0,...,t'\}; \tau^s=\{t'\}; \tau^i=\{t',...,T_f\}\setminus\tau^a$. Set $\tau^v=\emptyset$.\\
Else: &Set $\tau^v=\{t_0,...,T_f\}$; $\tau^a,\tau^s,\tau^i=\emptyset$.\\

\bottomrule\\
\end{tabular}
\caption{\label{tab:Umodule} Until module definition.}
\end{table}
\FloatBarrier

\begin{tcolorbox}[colback=gray!10!white,colframe=white!80!black,title=\color{black}{Definition \ref{FiniteTraceDef}: Formula Evaluation on Finite Traces (U)}]
LTL formula $\varphi$ is true on finite trace $\rho^{t_0...} = (L_{t_0},...,L_{T_f})$, denoted $\rho^{t_0...}\models\varphi$, if $T_0\leq t_0 \leq T_f$ \textbf{and}:
 \par$\rho^{t_0...}\models\varphi_1\mathbf{U}\varphi_2\quad$ \textbf{iff} $\quad \exists i\geq t_0$ s.t. $\rho^{i...}\models\varphi_2$ \textbf{and} $\rho^{k...} \models \varphi_1$ for all $k$ s.t. $t_0\leq k < i$ \textbf{and} $i\leq T_f$

\end{tcolorbox}

\begin{tcolorbox}[colback=gray!50!white,colframe=white!60!black,title=Properties of $\varphi$]
\par \textbf{Precondition}: $true$ (Def \ref{preconditionDef})
\par \textbf{Arbitrary suffixes}: $\rho^{t...}$ with $t'<t\leq T_f$, where $t'=\min\{i\ |\ \rho^{i...}\models\varphi_2\}$ (Def \ref{arbitraryDef})
\begin{itemize}
\item Consider $\varphi = \varphi_1\mathbf{U}\varphi_2$. By Def. \ref{FiniteTraceDef}, $\rho^{t_0...}\models\varphi$ iff there is some $i\geq t_0$ such that (1) $\rho^{i...}\models\varphi_2$ and (2) $\rho^{k...}\models\varphi_1$ for all $k$ s.t. $t_0\leq k < i$. This means that $\rho^{t_0...}\models\varphi$ depends on whether $\rho^{t...}\models\varphi_1$ or $\rho^{t...}\models\varphi_2$ for all $t$ from $t_0\leq t\leq i$.
\item Since existence of any such $i$ guarantees $\rho^{t_0...}\models\varphi$, we may select the minimum $i_{min}$ and note that $\rho^{t_0...}\models\varphi$ does not depend on $\rho^{t...}\models\varphi_1$ or $\rho^{t...}\models\varphi_2$ for any $i_{min}< t\leq T_f$. 

\item We claim now that, given $\rho^{t_0...}\models\varphi$, all arbitrary suffixes are found by taking all $\rho^{t...}$ where $t'=\min\{i\ |\ \rho^{i...}\models\varphi_2\}$, $t'<t\leq T_f$. Indeed, if $\rho^{t_0...}\models\varphi$, there must exist an $i_{min}$ where both $\rho^{i...}\models\varphi_2$ and $\rho^{k...}\models\varphi_1$ for all $k$ s.t. $t_0\leq k < i$. We have already shown that all $\rho^{t...}$ with $t>i$ are arbitrary. Suppose however that there is some `earlier' $t'$ (i.e., $t'<i_{min}$) where $\rho^{i...}\models\varphi_2$. Since $t'<i_{min}$, we must have $\rho^{k...}\models\varphi_1$ for all $k$ s.t. $t_0\leq k < t' < i$ as well, and thus $t'\in \{i\ |\ \rho^{i...}\models\varphi_2\}$.    

\item Thus, all $\rho^{t...}$ with $t'=\min\{i\ |\ \rho^{i...}\models\varphi_2\}$,  $t'<t\leq T_f$ are the arbitrary suffixes of $\varphi_1\mathbf{U}\varphi_2$.
\end{itemize}
\end{tcolorbox}

\begin{proof}[Proof: Soundness of Until Module] We examine the outputs of the Until module status by status. The status definition conditions are labeled \textbf{(1a)}, \textbf{(2a)}, etc. as in the reproduction of Definition \ref{activedef} above.

\begin{table}[h]
\centering
\begin{tabular}{lp{13.0 cm}}
\multicolumn{2}{l}{\textbf{Until Active status timeset $\tau^a$}}\\
\toprule
\multicolumn{2}{l}{(For $t'=t_0...T_f$: Load $\tau_1^q, \tau_2^q$ of $\varphi_1$ and $\varphi_2$ on $\rho^{t'...}$. Continue until $\tau_2^v = \emptyset$ or $\tau^v_1\neq\emptyset$ or $t'=T_f$.)}\\
\midrule
\textbf{If} $\tau_2^v = \emptyset$:&\textbf{Set} $\tau^a = \{t_0,...,t'\}$\\
\midrule
& $\bullet$ $\tau^v_2=\emptyset$ on $\rho^{t'...}$ iff $\rho^{t'...}\models\varphi_2$ by Def. \ref{activedef}. 
\par $\bullet$ $\tau^v_1=\emptyset$ on all $t< t'$ by the loop condition, so $\rho^{t...}\models\varphi_1$ for $t<t'$.
\par $\bullet$ Thus, a $t'$ exists s.t. $\rho^{t'...}\models \varphi_2$ and $\rho^{k...}\models\varphi_1$ for all $t_0\leq k < t'$ (see Def. \ref{FiniteTraceDef}) \textbf{(1a)} 
\par $\bullet$ $\rho^{t_0...}\models true$ \textbf{(2a)}
\par $\bullet$ Consider $\{i\ |\ \rho^{i...}\models\varphi_2\}$. When $t'$ equals any such $i$, we have $(\rho^{t'...}\models\varphi_2) \Rightarrow (\tau^v_2=\emptyset)$. Thus the loop terminates at any $t'=i$. 
\par $\bullet$ The loop strictly increases in $t'$. Thus $t'$ at termination satisfies $t'=\min\{i\ |\ \rho^{i...}\models\varphi_2\}$ in this case, and $\rho^{t...}$ is arbitrary for $t>t'$ \textbf{(3a for $t_0\leq t \leq t'$)}.
\par Therefore, $t_0\leq t \leq t'$ meet all active criteria and $\tau^a=\{t_0,...,t'\}$. \qed \\
\midrule
\textbf{Else}: &\textbf{Set} $\tau^a=\emptyset$.\\
\midrule
& $\bullet$ We have $\tau^v_2\neq \emptyset$ for all $t\leq t'$ (loop conditions), implying that $\rho^{t...}\not\models\varphi_2$ on all $t_0\leq t\leq t'$. Thus, no $t\leq t'$ is in $\{i\ |\ \rho^{i...}\models\varphi_2,\ \rho^{k...}\models\varphi_1\ \forall k\ s.t.\ t_0\leq k<i\}$. Therefore, the conditions for $\rho^{t_0...}\models\varphi_1\mathbf{U}\varphi_2$ are not met by any $t\leq t'$ by Def. \ref{FiniteTraceDef}.
\par $\bullet$ \textbf{If} $\tau^v_1\neq\emptyset$, we have $\rho^{t'...}\not\models\varphi_1$ (Def. \ref{activedef}). We see that for any $i> t'$, it is not true that $\rho^{k...} \models \varphi_1$ for all $k$ s.t. $t_0\leq k < i$ (simply select $k=t'$). Thus, there exists no $t'=i$ as required for $\rho^{t_0...}\models\varphi_1\mathbf{U}\varphi_2$ by Def. \ref{FiniteTraceDef}. 
\par $\bullet$ \textbf{If} $t'=T_f$, we see that $\rho^{t...}\not\models\varphi_2$ for all $t_0\leq t \leq T_f$. Once again, there is no $t'=i$ as required for $\rho^{t_0...}\models\varphi_1\mathbf{U}\varphi_2$ by Def. \ref{FiniteTraceDef}. 
\par Thus, $\rho^{t_0...}\not\models\varphi_1\mathbf{U}\varphi_2$ (1a is not met) and $\tau^a=\emptyset$.\qed 
\\
\bottomrule
\end{tabular}
\end{table}
\FloatBarrier

\begin{table}[h]
\centering
\begin{tabular}{lp{13.0 cm}}
\multicolumn{2}{l}{\textbf{Until Satisfied status timeset $\tau^s$}}\\
\toprule
\multicolumn{2}{l}{(For $t'=t_0...T_f$: Load $\tau_1^q, \tau_2^q$ of $\varphi_1$ and $\varphi_2$ on $\rho^{t'...}$. Continue until $\tau_2^v = \emptyset$ or $\tau^v_1\neq\emptyset$ or $t'=T_f$.)}\\
\midrule
\textbf{If} $\tau_2^v = \emptyset$:&\textbf{Set} $\tau^s=\{t'\}$\\
\midrule
& $\bullet$ $\varphi$ is active up to $t'$ \textbf{(1s for $t_0\leq t \leq t'$)}, \textbf{(2s for $t'$)}.
\par Therefore, \textbf{only} $t'$ meets all conditions at exit, and $\tau^s=\{t'\}$. \qed \\
\midrule
\textbf{Else}: &\textbf{Set} $\tau^s=\emptyset$.\\
\midrule
&Since $\varphi$ is not active at any $t$, (1s) is not met for any $t$, so $\tau^s=\emptyset$. \qed \\
\bottomrule
\end{tabular}
\end{table}
\FloatBarrier

\begin{table}[h]
\centering
\begin{tabular}{lp{13.0 cm}}
\multicolumn{2}{l}{\textbf{Until Inactive status timeset $\tau^i$}}\\
\toprule
\multicolumn{2}{l}{(For $t'=t_0...T_f$: Load $\tau_1^q, \tau_2^q$ of $\varphi_1$ and $\varphi_2$ on $\rho^{t'...}$. Continue until $\tau_2^v = \emptyset$ or $\tau^v_1\neq\emptyset$ or $t'=T_f$.)}\\
\midrule
\textbf{If} $\tau_2^v = \emptyset$:&\textbf{Set} $\tau^i=\{t',...,T_f\}\setminus\tau^a$\\
\midrule
&See Active proof of (1a) for \textbf{(1i)}. Then $\tau^i =\{t_0,...,T_f\}\setminus \tau^a$ (see \textbf{2i}). \qed  \\
\midrule
\textbf{Else}: &\textbf{Set} $\tau^i=\emptyset$\\
\midrule
&(1a) is not met iff (1i) is not met (see Active proof). Thus $\tau^i =\emptyset$. \qed \\
\bottomrule
\end{tabular}
\end{table}
\FloatBarrier

\begin{table}[h]
\centering
\begin{tabular}{lp{13.0 cm}}
\multicolumn{2}{l}{\textbf{Until Violated status timeset $\tau^v$}}\\
\toprule
\multicolumn{2}{l}{(For $t'=t_0...T_f$: Load $\tau_1^q, \tau_2^q$ of $\varphi_1$ and $\varphi_2$ on $\rho^{t'...}$. Continue until $\tau_2^v = \emptyset$ or $\tau^v_1\neq\emptyset$ or $t'=T_f$.)}\\
\midrule
\textbf{If} $\tau_2^v = \emptyset$:&\textbf{Set} $\tau^v=\emptyset$.\\
\midrule
& See proof of (1a) for $\rho^{t_0...}\models\varphi$. Thus \textbf{no} $t$ meets (1v) and $\tau^v=\emptyset$. \qed \\
\midrule
\textbf{Else}: &\textbf{Set} $\tau^v=\{t_0,...,T_f\}$\\
\midrule
&$\rho^{t_0...}\not\models\varphi$ by Active proof above \textbf{(1v)}. Thus $\tau^v=\{t_0,...,T_f\}$. \qed \\
\bottomrule
\end{tabular}
\end{table}
\FloatBarrier

\end{proof}


\subsection{Weak until}

\begin{table}[h]
\centering
\begin{tabular}{lp{13.0 cm}}
\toprule
 \textbf{W. until} $(W)$&LTL: $\varphi=\varphi_1\mathbf{W}\varphi_2$\\
 \midrule
For $t'=t_0...T_f$:& Load all $\tau_1^q, \tau_2^q$ of child nodes $\varphi_1$ and $\varphi_2$ on $\rho^{t'...}$. Continue until $\tau_2^v = \emptyset$ or $\tau^v_1\neq\emptyset$ or $t'=T_f$.\\
If $\tau_2^v = \emptyset$:&Set $\tau^a = \{t_0,...,t'\}; \tau^s=\{t'\}; \tau^i=\{t',...,T_f\}\setminus\tau^a$. Set $\tau^v=\emptyset$.\\
Else if $\tau^v_1\neq\emptyset$: &Set $\tau^v=\{t_0,...,T_f\}$; $\tau^a,\tau^s,\tau^i=\emptyset$.\\
Else: &Set $\tau^a=\{t_0,...,T_f\}; \tau^s=\{T_f\}, \tau^i = \emptyset$. Set $\tau^v=\emptyset$.\\

\bottomrule\\
\end{tabular}
\caption{\label{tab:Wmodule} Weak until module definition.}
\end{table}
\FloatBarrier

\begin{tcolorbox}[colback=gray!10!white,colframe=white!80!black,title=\color{black}{Definition \ref{FiniteTraceDef}: Formula Evaluation on Finite Traces (W)}]
LTL formula $\varphi$ is true on finite trace $\rho^{t_0...} = (L_{t_0},...,L_{T_f})$, denoted $\rho^{t_0...}\models\varphi$, if $T_0\leq t_0 \leq T_f$ \textbf{and}:
 \par$\rho^{t_0...} \models \varphi_1\mathbf{W}\varphi_2 \quad$ where $\quad \rho^{t_0...} \models (\varphi_1\mathbf{U}\varphi_2)\vee\mathbf{G}\varphi_1$

\begin{itemize}
    \item We have that $\rho^{t_0...}\models (\varphi_1\mathbf{U}\varphi_2) \vee \mathbf{G}\varphi_1$ if $\rho^{t_0...} \models \mathbf{G}\varphi_1$ \textbf{or} $\rho^{t_0...}\models  \varphi_1\mathbf{U}\varphi_2$. The latter condition is true iff $\exists i\geq t_0$ s.t. $\rho^{i...}\models\varphi_2$ and $\rho^{k...} \models \varphi_1$ for all $t_0\leq k < i$ and $i\leq T_f$ (by def. of $\mathbf{U}$). Thus,
\end{itemize}

 \par $\rho^{t_0...} \models \varphi_1\mathbf{W}\varphi_2 \quad$ \textbf{iff }$\quad (\exists i\geq t_0$ s.t. $\rho^{i...}\models\varphi_2$ \textbf{and} $\rho^{k...} \models \varphi_1$ for all $t_0\leq k < i$ \textbf{and} $i\leq T_f)$ \textbf{or} $(\rho^{i...}\models\mathbf{G}\varphi_1$ for all $i$ where $t_0\leq i\leq T_f)$

\end{tcolorbox}

\begin{tcolorbox}[colback=gray!50!white,colframe=white!60!black,title=Properties of $\varphi$]
\par \textbf{Precondition}: $true$ (Def \ref{preconditionDef})
\par \textbf{Arbitrary suffixes}: $\rho^{t...}$ with $t'<t\leq T_f$, where\\ $t'=\min\{i\ |\ \rho^{i...}\models\varphi_2\}$, or \textbf{none} if the set is empty (Def \ref{arbitraryDef})
\begin{itemize}  

\item By Def. \ref{arbitraryDef}, a suffix is arbitrary for $\varphi_1\mathbf{W}\varphi_2$ if the truth of $\rho^{t_0...}\models(\varphi_1\mathbf{U}\varphi_2)\vee\mathbf{G}\varphi_1$ (Def. \ref{FiniteTraceDef}) does not depend on $\rho^{t...}\models\varphi_1$ nor $\rho^{t...}\models\varphi_2$ at any $t$ for some $t'\leq t\leq T_f$. 

\item Taking $\psi_1 := \varphi_1\mathbf{U}\varphi_2$ and $\psi_2:=\mathbf{G}\varphi_1$, we may write  $\varphi = \psi_1\vee\psi_2$. Suppose now that $\rho^{t...}$ is arbitrary for $\psi_1$. By definition, this means that $\rho^{t_0...}\models\psi_1$ does not depend on $\rho^{t...}\models\varphi_1$ nor $\rho^{t...}\models\varphi_2$. Next, note that $(\rho\models\psi_1) \Rightarrow (\rho\models\psi_1\vee\psi_2)$. Thus, if $\rho^{t_0...}\models\psi_1$ is true regardless of $\rho^{t...}\models\varphi_1$ and $\rho^{t...}\models\varphi_2$, it follows that $\rho^{t...}\models\psi_1\vee\psi_2$ must be as well. Therefore, \textbf{since the suffix $\rho^{t...}$ is arbitrary for one member of the disjunction, it is arbitrary for the full disjunction as well.}

\item Thus, recalling the proof for Until, we see that suffixes become arbitrary via \\$\psi_1$ at $t'=\min\{i\ |\ \rho^{i...}\models\varphi_2\}$. For any such $t'$, all $\rho^{t...}$, $t\geq t'$ are arbitrary by Def. \ref{arbitraryDef}. Meanwhile, $\rho^{t_0...}\models\mathbf{G}\varphi_1$ has \textbf{no} arbitrary suffixes (see properties of Global operator above), so any arbitrary suffixes of the full disjunction $\varphi=\psi_1\vee\psi_2$ solely result from $\rho^{t'...}\models\psi_1$.

\item Thus, arbitrary suffixes will be $\rho^{t...}$ with $t'<t\leq T_f$, where\\ $t'=\min\{i\ |\ \rho^{i...}\models\varphi_2\}$ (\textbf{none} if the set is empty). 

\end{itemize}
\end{tcolorbox}

\begin{proof}[Proof: Soundness of Weak Until Module] We examine the outputs of the Weak Until module status by status. The status definition conditions are labeled \textbf{(1a)}, \textbf{(2a)}, etc. as in the reproduction of Definition \ref{activedef} above.

\begin{table}[h]
\centering
\begin{tabular}{lp{13.0 cm}}
\multicolumn{2}{l}{\textbf{Weak Until Active status timeset $\tau^a$}}\\
\toprule
\multicolumn{2}{l}{(For $t'=t_0...T_f$: Load $\tau_1^q, \tau_2^q$ of $\varphi_1$ and $\varphi_2$ on $\rho^{t'...}$. Continue until $\tau_2^v = \emptyset$ or $\tau^v_1\neq\emptyset$ or $t'=T_f$.)}\\
\midrule
\textbf{If} $\tau_2^v = \emptyset$:&\textbf{Set} $\tau^a = \{t_0,...,t'\}$\\
\midrule
& $\bullet$ $\tau^v_2=\emptyset$ on $\rho^{t'...}$ iff $\rho^{t'...}\models\varphi_2$ by Def. \ref{activedef}. 
\par $\bullet$ $\tau^v_1=\emptyset$ on all $t< t'$ by the loop condition, so $\rho^{t...}\models\varphi_1$ for $t<t'$.
\par $\bullet$ Thus, a $t'$ exists s.t. $\rho^{t'...}\models \varphi_2$ and $\rho^{k...}\models\varphi_1$ for all $t_0\leq k < t'$ (see Def. \ref{FiniteTraceDef}) \textbf{(1a)} 
\par $\bullet$ $\rho^{t_0...}\models true$ \textbf{(2a)}
\par $\bullet$ Consider $\{i\ |\ \rho^{i...}\models\varphi_2\}$. When $t'$ equals any such $i$, we have $(\rho^{t'...}\models\varphi_2) \Rightarrow (\tau^v_2=\emptyset)$. Thus the loop terminates at any $t'=i$. 
\par $\bullet$ The loop strictly increases in $t'$. Thus $t'$ at termination satisfies $t'=\min\{i\ |\ \rho^{i...}\models\varphi_2\}$ in this case, and $\rho^{t...}$ is arbitrary for $t>t'$ \textbf{(3a for $t_0\leq t \leq t'$)}.
\par Therefore, $t_0\leq t \leq t'$ meet all active criteria and $\tau^a=\{t_0,...,t'\}$. \qed \\

\midrule
\textbf{Else if} $\tau^v_1\neq\emptyset$: &\textbf{Set} $\tau^a=\emptyset$.\\
\midrule
& $\bullet$ Consider the disjunctive form $\varphi = (\varphi_1\mathbf{U}\varphi_2) \vee \mathbf{G}\varphi_1$. Take $\psi_1:=\varphi_1\mathbf{U}\varphi_2$, $\psi_2 := \mathbf{G}\varphi_1$. 
\par $\bullet$ We have $\tau^v_2\neq \emptyset$ for all $t\leq t'$ (loop conditions), implying that $\rho^{t...}\not\models\varphi_2$ on all $t_0\leq t\leq t'$. Thus, no $t\leq t'$ is in $\{i\ |\ \rho^{i...}\models\varphi_2,\ \rho^{k...}\models\varphi_1\ \forall k\ s.t.\ t_0\leq k<i\}$. Therefore, the conditions for $\rho^{t_0...}\models\psi_1$ are not met by any $t\leq t'$ by Def. \ref{FiniteTraceDef}.
\par $\bullet$ \textbf{Since} $\tau^v_1\neq\emptyset$, we have $\rho^{t'...}\not\models\varphi_1$ (Def. \ref{activedef}). Clearly $\rho^{t_0...}\not\models\psi_2$. Checking $\psi_1$, we see that for any $i> t'$, it is not true that $\rho^{k...} \models \varphi_1$ for all $k$ s.t. $t_0\leq k < i$ (simply select $k=t'$). We thus have $\rho^{t_0...}\not\models\psi_1$ \textbf{and} $\rho^{t_0...}\not\models\psi_2$.
\par Thus, $\rho^{t_0...}\not\models\varphi_1\mathbf{U}\varphi_2$ \textbf{and} $\rho^{t_0...}\not\models\mathbf{G}\varphi_1$, so $\rho^{t_0...}\not\models\varphi_1\mathbf{W}\varphi_2$ by Def. \ref{FiniteTraceDef} (1a is not met) and $\tau^a=\emptyset$.\qed \\
\midrule

\textbf{Else}: &\textbf{Set} $\tau^a=\{t_0,...,T_f\}$\\
\midrule
& $\bullet$ In this case, we have $t'=T_f$ with $\rho^{t'...}\models\varphi_1$ and $\rho^{t...}\not\models\varphi_2$ for all $t_0\leq t\leq T_f$. 
\par $\bullet$ We see that, in fact, $\rho^{t...}\models\varphi_1$ must then be true for all $t_0\leq t\leq T_f$ (loop conditions). Thus $\rho^{t_0...}\models\mathbf{G}\varphi_1$, which implies $\rho^{t_0...}\models\varphi_1\mathbf{W}\varphi_2$ \textbf{(1a)}.
\par $\bullet$ $\rho^{t_0...}\models true$ \textbf{(2a)}
\par $\bullet$ Since $\rho^{t...}\not\models\varphi_2$ for all $t_0\leq t\leq T_f$, $\{i\ |\ \rho^{i...}\models\varphi_2\} = \emptyset$. Reviewing the discussion of arbitrary suffixes for $\varphi_1\mathbf{W}\varphi_2$, this means that \textbf{no} arbitrary $\rho^{t...}$ exist \textbf{(3a for $t_0\leq t \leq T_f$)}.
\par Therefore, $t_0\leq t \leq t'$ meet all active criteria and $\tau^a=\{t_0,...,T_f\}$. \qed 
\\
\bottomrule
\end{tabular}
\end{table}
\FloatBarrier

\begin{table}[h]
\centering
\begin{tabular}{lp{13.0 cm}}
\multicolumn{2}{l}{\textbf{Weak Until Satisfied status timeset $\tau^s$}}\\
\toprule
\multicolumn{2}{l}{(For $t'=t_0...T_f$: Load $\tau_1^q, \tau_2^q$ of $\varphi_1$ and $\varphi_2$ on $\rho^{t'...}$. Continue until $\tau_2^v = \emptyset$ or $\tau^v_1\neq\emptyset$ or $t'=T_f$.)}\\
\midrule
\textbf{If} $\tau_2^v = \emptyset$:&\textbf{Set} $\tau^s=\{t'\}$\\
\midrule
& $\bullet$ $\varphi$ is active up to $t'$ \textbf{(1s for $t_0\leq t \leq t'$)}, \textbf{(2s for $t'$)}.
\par Therefore, \textbf{only} $t'$ meets all conditions at exit, and $\tau^s=\{t'\}$. \qed \\
\midrule

\textbf{Else if} $\tau^v_1\neq\emptyset$: &\textbf{Set} $\tau^s=\emptyset$.\\
\midrule
&Since $\varphi$ is not active at any $t$, (1s) is not met for any $t$, so $\tau^s=\emptyset$. \qed \\
\midrule

\textbf{Else}: &\textbf{Set} $\tau^s=\{T_f\}$\\
\midrule
&$\bullet$ $\varphi$ is active on all $t$ \textbf{(1s for $t_0\leq t \leq T_f$)}, \textbf{(2s for $T_f$)}.
\par Therefore, \textbf{only} $T_f$ meets all conditions at exit, and $\tau^s=\{T_f\}$. \qed \\
\bottomrule
\end{tabular}
\end{table}
\FloatBarrier

\begin{table}[h]
\centering
\begin{tabular}{lp{13.0 cm}}
\multicolumn{2}{l}{\textbf{Weak Until Inactive status timeset $\tau^i$}}\\
\toprule
\multicolumn{2}{l}{(For $t'=t_0...T_f$: Load $\tau_1^q, \tau_2^q$ of $\varphi_1$ and $\varphi_2$ on $\rho^{t'...}$. Continue until $\tau_2^v = \emptyset$ or $\tau^v_1\neq\emptyset$ or $t'=T_f$.)}\\
\midrule
\textbf{If} $\tau_2^v = \emptyset$:&\textbf{Set} $\tau^i=\{t',...,T_f\}\setminus\tau^a$\\
\midrule
&See Active proof of (1a) for \textbf{(1i)}. Then $\tau^i =\{t_0,...,T_f\}\setminus \tau^a$ (see \textbf{2i}). \qed  \\
\midrule

\textbf{Else if} $\tau^v_1\neq\emptyset$: &\textbf{Set} $\tau^i=\emptyset$\\
\midrule
&(1a) is not met iff (1i) is not met (see Active proof). Thus $\tau^i =\emptyset$. \qed \\
\midrule

\textbf{Else}: &\textbf{Set} $\tau^i = \emptyset$\\
\midrule
&$\varphi$ is active for all $t$ (2i not met for $t_0\leq t \leq T_f$). Thus $\tau^i=\emptyset$.\qed \\
\bottomrule
\end{tabular}
\end{table}
\FloatBarrier

\begin{table}[h]
\centering
\begin{tabular}{lp{13.0 cm}}
\multicolumn{2}{l}{\textbf{Weak Until Violated status timeset $\tau^v$}}\\
\toprule
\multicolumn{2}{l}{(For $t'=t_0...T_f$: Load $\tau_1^q, \tau_2^q$ of $\varphi_1$ and $\varphi_2$ on $\rho^{t'...}$. Continue until $\tau_2^v = \emptyset$ or $\tau^v_1\neq\emptyset$ or $t'=T_f$.)}\\
\midrule
\textbf{If} $\tau_2^v = \emptyset$:&\textbf{Set} $\tau^v=\emptyset$.\\
\midrule
& See proof of (1a) for $\rho^{t_0...}\models\varphi$. Thus \textbf{no} $t$ meets (1v) and $\tau^v=\emptyset$. \qed \\
\midrule

\textbf{Else if} $\tau^v_1\neq\emptyset$: &\textbf{Set} $\tau^v=\{t_0,...,T_f\}$\\
\midrule
&$\rho^{t_0...}\not\models\varphi$ by Active proof above \textbf{(1v)}. Thus $\tau^v=\{t_0,...,T_f\}$. \qed \\
\midrule

\textbf{Else}: &\textbf{Set} $\tau^v=\emptyset$\\
\midrule
&See proof of (1a) for $\rho^{t_0...}\models\varphi$. Thus \textbf{no} $t$ meets (1v) and $\tau^v=\emptyset$. \qed  \\
\bottomrule
\end{tabular}
\end{table}
\FloatBarrier

\end{proof}


\subsection{Strong release}

\begin{table}[h]
\centering
\begin{tabular}{lp{13.0 cm}}
\toprule
\textbf{S. release} $(M)$&LTL: $\varphi=\varphi_1\mathbf{M}\varphi_2$\\
 \midrule
 For $t'=t_0...T_f$:& Load all $\tau_1^q, \tau_2^q$ of child nodes $\varphi_1$ and $\varphi_2$ on $\rho^{t'...}$. Continue until $\tau_1^v = \emptyset$ or $\tau^v_2\neq\emptyset$ or $t'=T_f$.\\
If $\tau_2^v = \emptyset$ and $\tau_1^v = \emptyset$:&Set $\tau^a = \{t_0,...,t'\}; \tau^s=\{t'\}; \tau^i=\{t',...,T_f\}\setminus\tau^a$. Set $\tau^v=\emptyset$.\\
Else: &Set $\tau^v=\{t_0,...,T_f\}$; $\tau^a,\tau^s,\tau^i=\emptyset$.\\

\bottomrule\\
\end{tabular}
\caption{\label{tab:Mmodule} Strong release module definition.}
\end{table}
\FloatBarrier

\begin{tcolorbox}[colback=gray!10!white,colframe=white!80!black,title=\color{black}{Definition \ref{FiniteTraceDef}: Formula Evaluation on Finite Traces (M)}]
LTL formula $\varphi$ is true on finite trace $\rho^{t_0...} = (L_{t_0},...,L_{T_f})$, denoted $\rho^{t_0...}\models\varphi$, if $T_0\leq t_0 \leq T_f$ \textbf{and}:
 \par $\rho^{t_0...} \models \varphi_1\mathbf{M}\varphi_2 \quad$ where $\quad \rho^{t_0...} \models \varphi_2 \mathbf{U}(\varphi_1\wedge\varphi_2)$

\begin{itemize}
    \item We have that $\rho^{t_0...}\models \varphi_2 \mathbf{U}(\varphi_1\wedge\varphi_2)$ iff $\exists i,t_0\leq i\leq T_f$ s.t. $\rho^{i...}\models(\varphi_1 \wedge \varphi_2)$ and $\rho^{k...} \models \varphi_2$ for all $k$ s.t. $t_0\leq k < i)$ (by def. of $\mathbf{U}$). We can rewrite this:
\end{itemize}
 
 \par $\rho^{t_0...}\models\varphi_1 \mathbf{M} \varphi_2\quad$ \textbf{iff} $\quad\exists i,t_0\leq i\leq T_f$ s.t. $(\rho^{i...}\models \varphi_1) \wedge (\rho^{k...} \models \varphi_2$ for all $k$ s.t. $t_0\leq k \leq i)$

\end{tcolorbox}

\begin{tcolorbox}[colback=gray!50!white,colframe=white!60!black,title=Properties of $\varphi$]
\par \textbf{Precondition}: $true$ (Def \ref{preconditionDef})
\par \textbf{Arbitrary suffixes}: $\rho^{t...}$ with $t'<t\leq T_f$, where $t'=\min\{i\ |\ \rho^{i...}\models\varphi_1\}$ (Def \ref{arbitraryDef})
\begin{itemize}
\item Take $\psi_1:=\varphi_2$ and $\psi_2:=\varphi_1\wedge\varphi_2$, and consider $\varphi = \psi_1\mathbf{U}\psi_2$. From the discussion for the Until operator, we know the arbitrary suffixes for $\varphi=\psi_1\mathbf{U}\psi_2$ are $\rho^{t...}$ with $t'<t\leq T_f$ where $t'=\min\{i\ |\ \rho^{i...}\models\psi_2\}$. Reversing our substitution yields arbitrary $\rho^{t...}$ with $t'<t\leq T_f$ where $t'=\min\{i\ |\ \rho^{i...}\models(\varphi_1\wedge\varphi_2)\}$. 
\item However, we will now compare $t'=\min\{j\ |\ \rho^{j...}\models(\varphi_1\wedge\varphi_2)\}$ and $t''=\min\{j\ |\ \rho^{j...}\models\varphi_1\}$. First note that any $i$ as required by Def. \ref{FiniteTraceDef} must have $\rho^{i...}\models(\varphi_1\wedge\varphi_2)$. Thus, this $i\geq t'$. Next, note that $(\varphi_1\wedge\varphi_2)\Rightarrow \varphi_1$ logically, and thus $t''\leq t'$.
\item Suppose now that $t''<t'$. In this case, $\rho^{t''...}\models\varphi_1$ \textbf{and} $\rho^{t''...}\not\models\varphi_2$. Clearly $\rho^{k...}\not\models\varphi_2$ at $k=t''$, and thus it is not true that $\rho^{k...}\models\varphi_2$ $\forall k, t_0\leq k\leq t''$, which in turn means that $\rho^{k...}\models\varphi_2$ does not hold $\forall k, t_0\leq k\leq i$. The condition $t''<t'$ thus contradicts $\rho^{t_0...}\models\varphi$, and therefore $t'=t''$.
\item Since $t'=t''$, we may say that all $\rho^{t...}$ are arbitrary where $t'=\min\{i\ |\ \rho^{i...}\models\varphi_1\}$, $t'<t\leq T_f$. 
\end{itemize}
\end{tcolorbox}

\begin{proof}[Proof: Soundness of Strong Release Module] We examine the outputs of the Strong Release module status by status. The status definition conditions are labeled \textbf{(1a)}, \textbf{(2a)}, etc. as in the reproduction of Definition \ref{activedef} above.

\begin{table}[h]
\centering
\begin{tabular}{lp{13.0 cm}}
\multicolumn{2}{l}{\textbf{Strong Release Active status timeset $\tau^a$}}\\
\toprule
\multicolumn{2}{l}{(For $t'=t_0...T_f$: Load $\tau_1^q, \tau_2^q$ of $\varphi_1$ and $\varphi_2$ on $\rho^{t'...}$. Continue until $\tau_1^v = \emptyset$ or $\tau^v_2\neq\emptyset$ or $t'=T_f$.)}\\
\midrule
\textbf{If} $\tau_1^v,\tau_2^v = \emptyset$:&\textbf{Set} $\tau^a = \{t_0,...,t'\}$\\
\midrule
& $\bullet$ $\tau^v_1=\emptyset$ on $\rho^{t'...}$ iff $\rho^{t'...}\models\varphi_1$ and $\tau^v_2=\emptyset$ on $\rho^{t'...}$ iff $\rho^{t'...}\models\varphi_2$ by Def. \ref{activedef}. 
\par $\bullet$ $\tau^v_2=\emptyset$ on all $t< t'$ by the loop condition, so $\rho^{t...}\models\varphi_2$ for $t<t'$ as well.
\par $\bullet$ Thus, a $t'$ exists s.t. $\rho^{t'...}\models \varphi_1$ and $\rho^{k...}\models\varphi_2$ for all $t_0\leq k \leq t'$ (see Def. \ref{FiniteTraceDef}) \textbf{(1a)} 
\par $\bullet$ $\rho^{t_0...}\models true$ \textbf{(2a)}
\par $\bullet$ Consider $\{i\ |\ \rho^{i...}\models\varphi_1\}$. When $t'$ equals any such $i$, we have $(\rho^{t'...}\models\varphi_1) \Rightarrow (\tau^v_1=\emptyset)$ and the loop terminates. 
\par $\bullet$ The loop strictly increases in $t'$. Thus $t'$ at termination satisfies $t'=\min\{i\ |\ \rho^{i...}\models\varphi_1\}$ in this case, and $\rho^{t...}$ is arbitrary for $t>t'$ \textbf{(3a for $t_0\leq t \leq t'$)}.
\par Therefore, $t_0\leq t \leq t'$ meet all active criteria and $\tau^a=\{t_0,...,t'\}$. \qed \\
\midrule
\textbf{Else}: &\textbf{Set} $\tau^a=\emptyset$.\\
\midrule
& $\bullet$ Firstly, $\tau^v_1\neq \emptyset$ for all $t< t'$ (loop conditions), implying that $\rho^{t...}\not\models\varphi_1$ on all $t_0\leq t\leq t'$. Thus, no $t< t'$ is in $\{i\ |\ \rho^{i...}\models\varphi_1,\ \rho^{k...}\models\varphi_2\ \forall k\ s.t.\ t_0\leq k\leq i\}$. Therefore, the conditions for $\rho^{t_0...}\models\varphi_1\mathbf{M}\varphi_2$ are not met by any $t< t'$ by Def. \ref{FiniteTraceDef}.
\par $\bullet$ \textbf{If} $\tau^v_1\neq\emptyset$ at $t'$ as well, then termination occurred either because (1) $\tau^v_2\neq\emptyset$ or (2) $t'=T_f$. 
\par $\bullet$ In case (1), we have $\rho^{t'...}\not\models\varphi_2$ (Def. \ref{activedef}), and see that for any $i\geq t'$, it is not true that $\rho^{k...} \models \varphi_2$ for all $k$ s.t. $t_0\leq k \leq i$ (simply select $k=t'$). Thus, there exists no $t'=i$ as required for $\rho^{t_0...}\models\varphi_1\mathbf{M}\varphi_2$ by Def. \ref{FiniteTraceDef}. 
\par $\bullet$ In case (2), $t'=T_f$. If $\rho^{t'...}\not\models\varphi_2$ additionally, see case (1); otherwise,  $\rho^{t'...}\models\varphi_2$. Here, see that $\rho^{t...}\not\models\varphi_1$ for all $t_0\leq t \leq T_f$ (loop conditions); once again, there exists no $t'=i$ as required for $\rho^{t_0...}\models\varphi_1\mathbf{M}\varphi_2$ by Def. \ref{FiniteTraceDef}. 
\par $\bullet$ \textbf{If} $\tau^v_1=\emptyset$ at $t'$, then $\tau^v_2$ must be nonempty. Thus, $\rho^{t'...}\models\varphi_1$ and $\rho^{t'...}\not\models\varphi_2$. Similarly to case (1) above, it must not be true that $\rho^{k...}\models\varphi_2$ for all $k$ s.t. $t_0\leq k \leq i$ for any selection of $i\geq t'$. Once more, there exists no $t'=i$ as required for $\rho^{t_0...}\models\varphi_1\mathbf{M}\varphi_2$ by Def. \ref{FiniteTraceDef}.
\par Thus, $\rho^{t_0...}\not\models\varphi_1\mathbf{M}\varphi_2$ (1a is not met) and $\tau^a=\emptyset$.\qed 
\\
\bottomrule
\end{tabular}
\end{table}
\FloatBarrier

\begin{table}[h]
\centering
\begin{tabular}{lp{13.0 cm}}
\multicolumn{2}{l}{\textbf{Strong Release Satisfied status timeset $\tau^s$}}\\
\toprule
\multicolumn{2}{l}{(For $t'=t_0...T_f$: Load $\tau_1^q, \tau_2^q$ of $\varphi_1$ and $\varphi_2$ on $\rho^{t'...}$. Continue until $\tau_1^v = \emptyset$ or $\tau^v_2\neq\emptyset$ or $t'=T_f$.)}\\
\midrule
\textbf{If} $\tau_1^v,\tau_2^v = \emptyset$:&\textbf{Set} $\tau^s=\{t'\}$\\
\midrule
& $\bullet$ $\varphi$ is active up to $t'$ \textbf{(1s for $t_0\leq t \leq t'$)}, \textbf{(2s for $t'$)}.
\par Therefore, \textbf{only} $t'$ meets all conditions at exit, and $\tau^s=\{t'\}$. \qed \\
\midrule
\textbf{Else}: &\textbf{Set} $\tau^s=\emptyset$.\\
\midrule
&Since $\varphi$ is not active at any $t$, (1s) is not met for any $t$, so $\tau^s=\emptyset$. \qed \\
\bottomrule
\end{tabular}
\end{table}
\FloatBarrier

\begin{table}[h]
\centering
\begin{tabular}{lp{13.0 cm}}
\multicolumn{2}{l}{\textbf{Strong Release Inactive status timeset $\tau^i$}}\\
\toprule
\multicolumn{2}{l}{(For $t'=t_0...T_f$: Load $\tau_1^q, \tau_2^q$ of $\varphi_1$ and $\varphi_2$ on $\rho^{t'...}$. Continue until $\tau_1^v = \emptyset$ or $\tau^v_2\neq\emptyset$ or $t'=T_f$.)}\\
\midrule
\textbf{If} $\tau_1^v,\tau_2^v = \emptyset$:&\textbf{Set} $\tau^i=\{t',...,T_f\}\setminus\tau^a$\\
\midrule
&See Active proof of (1a) for \textbf{(1i)}. Then $\tau^i =\{t_0,...,T_f\}\setminus \tau^a$ (see \textbf{2i}). \qed  \\
\midrule
\textbf{Else}: &\textbf{Set} $\tau^i=\emptyset$\\
\midrule
&(1a) is not met iff (1i) is not met (see Active proof). Thus $\tau^i =\emptyset$. \qed \\
\bottomrule
\end{tabular}
\end{table}
\FloatBarrier

\begin{table}[h]
\centering
\begin{tabular}{lp{13.0 cm}}
\multicolumn{2}{l}{\textbf{Strong Release Violated status timeset $\tau^v$}}\\
\toprule
\multicolumn{2}{l}{(For $t'=t_0...T_f$: Load $\tau_1^q, \tau_2^q$ of $\varphi_1$ and $\varphi_2$ on $\rho^{t'...}$. Continue until $\tau_1^v = \emptyset$ or $\tau^v_2\neq\emptyset$ or $t'=T_f$.)}\\
\midrule
\textbf{If} $\tau_1^v,\tau_2^v = \emptyset$:&\textbf{Set} $\tau^v=\emptyset$.\\
\midrule
& See proof of (1a) for $\rho^{t_0...}\models\varphi$. Thus \textbf{no} $t$ meets (1v) and $\tau^v=\emptyset$. \qed \\
\midrule
\textbf{Else}: &\textbf{Set} $\tau^v=\{t_0,...,T_f\}$\\
\midrule
&$\rho^{t_0...}\not\models\varphi$ by Active proof above \textbf{(1v)}. Thus $\tau^v=\{t_0,...,T_f\}$. \qed \\
\bottomrule
\end{tabular}
\end{table}
\FloatBarrier

\end{proof}

\subsection{Release}

\begin{table}[h]
\centering
\begin{tabular}{lp{13.0 cm}}
\toprule
 \textbf{Release} $(R)$&LTL: $\varphi=\varphi_1\mathbf{R}\varphi_2$\\
 \midrule
For $t'=t_0...T_f$:& Load all $\tau_1^q, \tau_2^q$ of child nodes $\varphi_1$ and $\varphi_2$ on $\rho^{t'...}$. Continue until $\tau_1^v = \emptyset$ or $\tau^v_2\neq\emptyset$ or $t'=T_f$.\\
If $\tau_1^v,\tau_2^v = \emptyset$:&Set $\tau^a = \{t_0,...,t'\}; \tau^s=\{t'\}; \tau^i=\{t',...,T_f\}\setminus\tau^a$. Set $\tau^v=\emptyset$.\\
Else if $\tau^v_2\neq\emptyset$: &Set $\tau^v=\{t_0,...,T_f\}$; $\tau^a,\tau^s,\tau^i=\emptyset$.\\
Else: &Set $\tau^a=\{t_0,...,T_f\}; \tau^s=\{T_f\}, \tau^i = \emptyset$. Set $\tau^v=\emptyset$.\\

\bottomrule\\
\end{tabular}
\caption{\label{tab:Rmodule} Release module definition.}
\end{table}
\FloatBarrier

\begin{tcolorbox}[colback=gray!10!white,colframe=white!80!black,title=\color{black}{Definition \ref{FiniteTraceDef}: Formula Evaluation on Finite Traces (R)}]
LTL formula $\varphi$ is true on finite trace $\rho^{t_0...} = (L_{t_0},...,L_{T_f})$, denoted $\rho^{t_0...}\models\varphi$, if $T_0\leq t_0 \leq T_f$ \textbf{and}:
 \par$\rho^{t_0...} \models \varphi_1\mathbf{R}\varphi_2 \quad$ where $\quad \rho^{t_0...} \models \neg(\neg\varphi_1\mathbf{U}\neg\varphi_2)$

 \begin{itemize}
     \item Consider $\neg(\psi_1\mathbf{U}\psi_2)$. This is true on $\rho^{t_0...}$ iff $\neg(\exists i\geq t_0$ s.t. $\rho^{i...}\models\psi_2$ and $\rho^{k...} \models \psi_1$ for all $k$ s.t. $t_0\leq k < i)$ (by def. of $\mathbf{U}$) and $i\leq T_f$. The former is equivalent to requiring $\not\exists i\geq t_0$ s.t. ($\rho^{i...}\models\psi_2$ and $\rho^{k...} \models \psi_1$ for all $i,k$ where $t_0\leq k < i$). 
     \item This means that for all $i$ with $\rho^{i...}\models\psi_2$ (if such $i$ exist), there exists some $k$ with $t_0\leq k < i$ such that $\rho^{k...}\not\models\psi_1$. More formally, $\neg(\psi_1\mathbf{U}\psi_2)$ iff for all $i$ s.t. $t_0\leq i\leq T_f$, $\neg(\rho^{i...}\models\psi_2)\vee (\exists k<i$ s.t. $\rho^{k...}\not\models\psi_1)$.
     \item Substitute $\psi_1 := \neg\varphi_1$ and $\psi_2 := \neg\varphi_2$. Then $\rho^{t_0...}\models\neg(\neg\varphi_1 \mathbf{U} \neg\varphi_2)$ iff for all $i$ s.t. $t_0\leq i\leq T_f$, $\neg(\rho^{i...}\not\models\varphi_2)\vee (\exists k<i$ s.t. $\rho^{k...}\models\varphi_1)$, which yields
 \end{itemize}

 \par$\rho^{t_0...}\models\varphi_1 \mathbf{R} \varphi_2\quad$ \textbf{iff} $\quad$ for all $i$ s.t. $t_0\leq i\leq T_f$, $(\rho^{i...}\models\varphi_2)\vee (\exists k<i$ s.t. $\rho^{k...}\models\varphi_1)$

\end{tcolorbox}

\begin{tcolorbox}[colback=gray!50!white,colframe=white!60!black,title=Properties of $\varphi$]
\par \textbf{Precondition}: $true$ (Def \ref{preconditionDef})
\par \textbf{Arbitrary suffixes}: $\rho^{t...}$ with $t'<t\leq T_f$, where $t'=\min\{k\ |\ \rho^{k...}\models\varphi_1\}$, or \textbf{none} if the set is empty (Def \ref{arbitraryDef})
\begin{itemize}
\item Consider $t'=\min\{k\ |\ \rho^{k...}\models\varphi_1\}$. First, suppose $t'$ exists. Then, for all $t>t'$, there exists $k<t$ s.t. $\rho^{k...}\models\varphi_1$, and thus $\rho^{t_0...}\models\varphi$ by Def. \ref{FiniteTraceDef} regardless of $\rho^{t...}\models\varphi_1$ or  $\rho^{t...}\models\varphi_2$. Therefore, all such $\rho^{t...},t>t'$ are arbitrary. On the other hand, for $t\leq t'$, there must not exist such a $k$, since $t'$ is indeed the minimum. Thus, for $\rho^{t_0...}\models\varphi$, we must have $\rho^{t...}\models\varphi_2$, meaning that no $t\leq t'$ gives an arbitrary $\rho^{t...}$ by definition.
\item Now suppose $t'$ does not exist. Then there is no $k$ s.t. $\rho^{k...}\models\varphi_1$, and thus $\rho^{t...}\models\varphi_2$ is necessary at all $t$ for $\rho^{t_0...}\models\varphi$. This results in no arbitrary suffixes.
\item Thus, arbitrary suffixes will be all $\rho^{t...}$ with $t'<t\leq T_f$, where\\ $t'=\min\{i\ |\ \rho^{i...}\models\varphi_1\}$ (\textbf{none} if the set is empty). 
\end{itemize}
\end{tcolorbox}

\begin{proof}[Proof: Soundness of Release Module] We examine the outputs of the Release module status by status. The status definition conditions are labeled \textbf{(1a)}, \textbf{(2a)}, etc. as in the reproduction of Definition \ref{activedef} above.

\begin{table}[h]
\centering
\begin{tabular}{lp{13.0 cm}}
\multicolumn{2}{l}{\textbf{Release Active status timeset $\tau^a$}}\\
\toprule
\multicolumn{2}{l}{(For $t'=t_0...T_f$: Load $\tau_1^q, \tau_2^q$ of $\varphi_1$ and $\varphi_2$ on $\rho^{t'...}$. Continue until $\tau_1^v = \emptyset$ or $\tau^v_2\neq\emptyset$ or $t'=T_f$.)}\\
\midrule
\textbf{If} $\tau_1^v,\tau_2^v = \emptyset$:&\textbf{Set} $\tau^a = \{t_0,...,t'\}$\\
\midrule
& $\bullet$ $\tau^v_1=\emptyset$ on $\rho^{t'...}$ iff $\rho^{t'...}\models\varphi_1$ and $\tau^v_2=\emptyset$ on $\rho^{t'...}$ iff $\rho^{t'...}\models\varphi_2$ by Def. \ref{activedef}. 
\par $\bullet$ $\tau^v_2=\emptyset$ on all $t< t'$ by the loop condition, so $\rho^{t...}\models\varphi_2$ for $t<t'$ as well.
\par $\bullet$ Thus, a $t'$ exists s.t. $\rho^{t'...}\models \varphi_1$ and $\rho^{k...}\models\varphi_2$ for all $t_0\leq k \leq t'$ (see Def. \ref{FiniteTraceDef}) \textbf{(1a)} 
\par $\bullet$ $\rho^{t_0...}\models true$ \textbf{(2a)}
\par $\bullet$ Consider $\{i\ |\ \rho^{i...}\models\varphi_1\}$. When $t'$ equals any such $i$, we have $(\rho^{t'...}\models\varphi_1) \Rightarrow (\tau^v_1=\emptyset)$. Thus the loop terminates at any $t'=i$. 
\par $\bullet$ The loop strictly increases in $t'$. Thus $t'$ at termination satisfies $t'=\min\{i\ |\ \rho^{i...}\models\varphi_1\}$ in this case, and $\rho^{t...}$ is arbitrary for $t>t'$ \textbf{(3a for $t_0\leq t \leq t'$)}.
\par Therefore, $t_0\leq t \leq t'$ meet all active criteria and $\tau^a=\{t_0,...,t'\}$. \qed \\
\midrule
\textbf{Else if} $\tau^v_2\neq\emptyset$: &\textbf{Set} $\tau^a=\emptyset$.\\
\midrule
& $\bullet$ For $\rho^{t_0...}\models\varphi_1\mathbf{R}\varphi_2$, at all $t$ we must have either $\rho^{t...}\models\varphi_2$ or $\exists k<t$ s.t. $\rho^{k...}\models\varphi_1$.
\par $\bullet$ Here, $\tau^v_2\neq\emptyset$, and thus $\rho^{t...}\not\models\varphi_2$. Additionally, we observe that $\rho^{k...}\not\models\varphi_1$ for any $k<t$ (since $\tau^v_1=\emptyset$ at such a $k$, and the loop would have terminated previously). 
\par Therefore, $\rho^{t_0...}\not\models\varphi_1\mathbf{R}\varphi_2$ by Def. \ref{FiniteTraceDef} (1a is not met) and $\tau^a=\emptyset$.\qed \\
\midrule

\textbf{Else}: &\textbf{Set} $\tau^a=\{t_0,...,T_f\}$\\
\midrule
& $\bullet$ In this case, we have  $\rho^{t'...}\models\varphi_2$ and $\rho^{t...}\not\models\varphi_1$, so termination must have occurred due to $t=T_f$. 
\par $\bullet$ Moreover, this means that $\rho^{t'...}\models\varphi_2$ for all $t_0\leq t\leq T_f$ (or the loop would have terminated previously) \textbf{(1a)}.
\par $\bullet$ $\rho^{t_0...}\models true$ \textbf{(2a)}
\par $\bullet$ Since $\rho^{t...}\not\models\varphi_1$ for all $t_0\leq t\leq T_f$, $\{i\ |\ \rho^{i...}\models\varphi_1\} = \emptyset$. This means that \textbf{no} arbitrary $\rho^{t...}$ exist \textbf{(3a for $t_0\leq t \leq T_f$)}.
\par Therefore, $t_0\leq t \leq t'$ meet all active criteria and $\tau^a=\{t_0,...,T_f\}$. \qed 
\\
\bottomrule
\end{tabular}
\end{table}
\FloatBarrier

\begin{table}[h]
\centering
\begin{tabular}{lp{13.0 cm}}
\multicolumn{2}{l}{\textbf{Release Satisfied status timeset $\tau^s$}}\\
\toprule
\multicolumn{2}{l}{(For $t'=t_0...T_f$: Load $\tau_1^q, \tau_2^q$ of $\varphi_1$ and $\varphi_2$ on $\rho^{t'...}$. Continue until $\tau_1^v = \emptyset$ or $\tau^v_2\neq\emptyset$ or $t'=T_f$.)}\\
\midrule
\textbf{If} $\tau_1^v,\tau_2^v = \emptyset$:&\textbf{Set} $\tau^s=\{t'\}$\\
\midrule
& $\bullet$ $\varphi$ is active up to $t'$ \textbf{(1s for $t_0\leq t \leq t'$)}, \textbf{(2s for $t'$)}.
\par Therefore, \textbf{only} $t'$ meets all conditions at exit, and $\tau^s=\{t'\}$. \qed \\
\midrule
\textbf{Else if} $\tau^v_2\neq\emptyset$: &\textbf{Set} $\tau^s=\emptyset$.\\
\midrule
&Since $\varphi$ is not active at any $t$, (1s) is not met for any $t$, so $\tau^s=\emptyset$. \qed \\
\midrule

\textbf{Else}: &\textbf{Set} $\tau^s=\{T_f\}$\\
\midrule
&$\bullet$ $\varphi$ is active on all $t$ \textbf{(1s for $t_0\leq t \leq T_f$)}, \textbf{(2s for $T_f$)}.
\par Therefore, \textbf{only} $T_f$ meets all conditions at exit, and $\tau^s=\{T_f\}$. \qed \\
\bottomrule
\end{tabular}
\end{table}
\FloatBarrier

\begin{table}[h]
\centering
\begin{tabular}{lp{13.0 cm}}
\multicolumn{2}{l}{\textbf{Release Inactive status timeset $\tau^i$}}\\
\toprule
\multicolumn{2}{l}{(For $t'=t_0...T_f$: Load $\tau_1^q, \tau_2^q$ of $\varphi_1$ and $\varphi_2$ on $\rho^{t'...}$. Continue until $\tau_1^v = \emptyset$ or $\tau^v_2\neq\emptyset$ or $t'=T_f$.)}\\
\midrule
\textbf{If} $\tau_1^v,\tau_2^v = \emptyset$:&\textbf{Set} $\tau^i=\{t',...,T_f\}\setminus\tau^a$\\
\midrule
&See Active proof of (1a) for \textbf{(1i)}. Then $\tau^i =\{t_0,...,T_f\}\setminus \tau^a$ (see \textbf{2i}). \qed  \\
\midrule
\textbf{Else if} $\tau^v_2\neq\emptyset$: &\textbf{Set} $\tau^i=\emptyset$\\
\midrule
&(1a) is not met iff (1i) is not met (see Active proof). Thus $\tau^i =\emptyset$. \qed \\
\midrule

\textbf{Else}: &\textbf{Set} $\tau^i = \emptyset$\\
\midrule
&$\varphi$ is active for all $t$ (2i not met for $t_0\leq t \leq T_f$). Thus $\tau^i=\emptyset$.\qed \\
\bottomrule
\end{tabular}
\end{table}
\FloatBarrier

\begin{table}[h]
\centering
\begin{tabular}{lp{13.0 cm}}
\multicolumn{2}{l}{\textbf{Release Violated status timeset $\tau^v$}}\\
\toprule
\multicolumn{2}{l}{(For $t'=t_0...T_f$: Load $\tau_1^q, \tau_2^q$ of $\varphi_1$ and $\varphi_2$ on $\rho^{t'...}$. Continue until $\tau_1^v = \emptyset$ or $\tau^v_2\neq\emptyset$ or $t'=T_f$.)}\\
\midrule
\textbf{If} $\tau_1^v,\tau_2^v = \emptyset$:&\textbf{Set} $\tau^v=\emptyset$.\\
\midrule
& See proof of (1a) for $\rho^{t_0...}\models\varphi$. Thus \textbf{no} $t$ meets (1v) and $\tau^v=\emptyset$. \qed \\
\midrule
\textbf{Else if} $\tau^v_2\neq\emptyset$: &\textbf{Set} $\tau^v=\{t_0,...,T_f\}$\\
\midrule
&$\rho^{t_0...}\not\models\varphi$ by Active proof above \textbf{(1v)}. Thus $\tau^v=\{t_0,...,T_f\}$. \qed \\
\midrule

\textbf{Else}: &\textbf{Set} $\tau^v=\emptyset$\\
\midrule
&See proof of (1a) for $\rho^{t_0...}\models\varphi$. Thus \textbf{no} $t$ meets (1v) and $\tau^v=\emptyset$. \qed  \\
\bottomrule
\end{tabular}
\end{table}
\FloatBarrier

\end{proof}
\end{document}